\documentclass[lettersize,journal]{IEEEtran}

\usepackage{times}
\usepackage{setspace}
\usepackage{url}
\spacing{1}
\usepackage[utf8]{inputenc}
\usepackage[T1]{fontenc}
\usepackage{graphicx}		% For \includegraphics
\usepackage{wrapfig}
\usepackage[format=plain,font=footnotesize,labelfont=bf,labelsep=period]{caption}
\usepackage{sidecap} % For sidecaptions for figures
\usepackage{subfig}
\usepackage[export]{adjustbox}
\usepackage[font=small]{caption}
\usepackage{float}

\usepackage{amsmath} % assumes amsmath package installed
\usepackage{amssymb}  % assumes amsmath package installed
\usepackage{amsthm}
\usepackage{mathtools}
\usepackage[normalem]{ulem}
\usepackage{paralist}	% For enumerations with better titled numbers
\usepackage[space]{grffile} % Space in filenames
\usepackage{color}
\usepackage{enumitem}
\usepackage{bm}

\theoremstyle{definition}
\newtheorem{definition}{Definition}
\theoremstyle{remark}

\theoremstyle{definition}
\newtheorem{assumption}{Assumption}
\theoremstyle{definition}

\newcommand{\R}{\mathbb{R}}
\newcommand{\C}{\mathcal{C}}

\newcommand{\K}{\mathcal{K}}

\definecolor{darkblue}{RGB}{0,0,102}
\definecolor{lightblue}{RGB}{77,77,148}

\definecolor{gold}{RGB}{234, 170, 0}
\definecolor{metallic_gold}{RGB}{139, 111, 78}

\newcommand{\der}[2]{\frac{\mathrm{d} #1 }{\mathrm{d} #2 }}
\newcommand{\derp}[2]{\frac{\partial #1 }{\partial #2 }}

\DeclareMathOperator*{\argmin}{argmin}

\usepackage{xfrac}

\usepackage{amsmath,amsfonts}
\usepackage{algorithmic}
\usepackage{algorithm}
\usepackage{array}
\usepackage{textcomp}
\usepackage{stfloats}
\usepackage{url}
\usepackage{verbatim}
\usepackage{graphicx}
\usepackage{cite}
\usepackage{cases}
\usepackage{empheq}
\usepackage{makecell}
\usepackage{multirow}
\begin{document}

\title{Experimental Validation of a Safe 
Controller Integration Scheme for Connected Automated Trucks}

\author{Anil Alan$^{1}$, Chaozhe R. He$^{1,2}$, Tamas G. Molnar$^{3}$, Johaan C. Mathew$^{4}$, A. Harvey Bell$^{5}$, and G\'abor Orosz$^{1,6}$

\thanks{This research is supported by the Navistar, Inc.}
\thanks{$^{1}$A. Alan, C. R. He, and G. Orosz are with the Department of Mechanical Engineering, University of Michigan, Ann Arbor, MI 48109, USA, ${\tt\small \{anilalan, orosz\}@umich.edu}$}%
\thanks{$^{2}$C. R. He is also with {Plus.ai Inc.}, Santa Clara, CA 95014, USA, 
${\tt\small chaozhe.he@plus.ai}$  }
\thanks{$^{3}$T. G. Molnar is with the Department of Mechanical and Civil Engineering, California Institute of Technology, Pasadena, CA 91125, USA, ${\tt\small tmolnar@caltech.edu}$}%
\thanks{$^{4}$J. C. Mathew is with Visteon Corporation, Van Buren Township, MI 48111, USA,
${\tt\small jmathew2@visteon.com }$  }
\thanks{$^{5}$A. Harvey Bell is with Multidisciplinary Design Program, University of Michigan, Ann Arbor, MI 48109, USA,
${\tt\small ahbelliv@umich.edu}$  }
\thanks{$^{6}$G. Orosz is also with the Department of Civil and Environmental Engineering, University of Michigan, Ann Arbor, MI 48109, USA}%
}

\maketitle
\thispagestyle{empty}

\begin{abstract}
Accomplishing safe and efficient driving is one of the predominant challenges in the controller design of connected automated vehicles (CAVs).
It is often more convenient to address these goals separately and integrate the resulting controllers.
In this study, we propose a controller integration scheme to fuse performance-based controllers and safety-oriented controllers 
safely for the longitudinal motion of a CAV. 
The resulting structure is compatible with a large class of controllers, and offers flexibility to design each controller individually without affecting the performance of the others.  
We implement the proposed safe integration scheme on a connected automated truck using an optimal-in-energy controller and a safety-oriented connected cruise controller.
We validate the premise of the safe integration through
experiments with a full-scale truck in
two scenarios: a controlled experiment on a test track  and a real-world experiment on a public highway.
In both scenarios, 
we achieve energy-efficient driving
without violating safety.
\end{abstract}


%%%%%%%%%%%%%%%%%%%%%%%%%%%%%%%%%%%%%%%%%%%%%%%%%%%%%%%%%%%%%%%%%%%%%%%%%%%%%%%%%
\section{Introduction}

The rapid progress in automated vehicle (AV) technology is projected to lead to considerable amount of AVs on public roads in the foreseeable future, even with conservative estimates \cite{litman2020autonomous}. 
AVs are expected to bring prospects to individuals and society, including improved mobility, comfort, energy and time efficiency, and a reduction in carbon emission \cite{chan2017advancements,brown2014analysis}. 
While each of these prospects poses essential objectives to be optimized in AV design, safety is yet to remain the most critical requirement. Indeed, studies show that 93\% of the total traffic accidents per year in the US are caused by human-related errors \cite{NationalReport:08}, which is a factor that can potentially be reduced or diminished with reliable AV technologies \cite{morando2018studying}. 

Safety in commercially available AVs is typically maintained with features such as automatic emergency braking \cite{cicchino17effectiveness} and lane keeping \cite{xu2019design}, where the longitudinal and lateral motion of the vehicle is controlled, respectively, based on the information from perception systems.
A wide selection of onboard range sensors can be utilized for detection purposes, such as radar, lidar, and camera \cite{yurtsever2020survey}. 
Additionally, recent advancements in communication technology have paved the way for wireless connectivity that provide reliable information exchange between different road users and the infrastructure, which, when integrated with AV technology, leads to connected automated vehicle (CAV) technology.
Connectivity bears a substantial potential to amplify the above-mentioned prospects of AVs \cite{talebpour2016influence,vahidi2018energy}, and some of these improvements have been reported in earlier experimental studies \cite{alam2015heavy,ge2018experimental,pelletier2021vehicle}.

It is crucial to employ the right controller strategies to take full advantage of the benefits offered by the CAV technology. 
The control problem for a CAV usually consists of two main goals: maximizing the outcome of one or a combination of desired prospects while keeping the system safe.
The first goal is typically constructed in the optimization context, such as finding the optimal controller parameters of a given controller using classical linear control techniques \cite{xu2019design} and data-driven methods \cite{he2019fuel}, or finding optimal trajectories for a finite horizon as in model predictive control (MPC) \cite{li2017performance,turri2017model,HomChaudhuri2017FastMPC,liu2018nonlinear}. 
MPC may handle safety requirements through constraints to address the second goal. However, its computational burden for solving complex nonlinear optimization problems in a rolling horizon fashion makes it challenging to be implemented into an on-board unit with limited computation capacity. 

An alternative approach to satisfy both control tasks (performance and safety) is to focus on each problem separately and then integrate the resulting controller strategies in the implementation. 
Integration methods typically operate in the scheme of correcting (or interfering with) a performance-based controller according to a specific safety task.
The study \cite{magdici2017adaptive} offers an adaptive cruise control (ACC) for longitudinal control with a switching structure; the nominal controller is switched to a braking trajectory when its output no longer satisfies a prescribed minimum safe distance from the preceding vehicle.
In \cite{groelke2021predictive}, a command/reference governer method is utilized to adjust an existing higher-level controller to satisfy safety constraints over a horizon. 
Another scheme is proposed in \cite{nilsson2015correct}, where formal methods guarantee safety specifications with a correct-by-construction controller. This controller can be used as a supervisor to interfere with an existing controller. 
In \cite{althoff2014online}, an online safety verification framework is proposed for a given reference motion trajectory; the reachable set of states of the AV and other vehicles are calculated considering the input limitations and disturbances in the system.
The framework allows only the safe intended plans among the newly calculated plans in case a given trajectory plan is deemed unsafe.
Although shown effective, the resulting controller structures in these existing solutions lack the flexibility to employ  
\emph{any} existing safety-oriented controller that might have been exhaustively tested on real road conditions.
Such as the controllers running on commercially available AVs or controllers that have been evaluated as safe using safety assessment methods \cite{riedmaier2020survey}.

We have two main contributions to this study.
Our first contribution is to create a simple integration scheme to combine one or multiple performance-based controllers with safety-oriented controllers of our choice to achieve optimal performance and ensure safety in longitudinal car-following scenarios involving a CAV. 
The resulting scheme should be flexible and compatible with any safety-oriented controller.
The proposed structure, which is based on control barrier functions (CBFs), ensures that the system is \emph{at least} as safe as the employed safety-oriented controller makes it to be. 
CBFs have been used to synthesize safe-by-construction controllers \cite{ames2017control} and have been implemented on a comprehensive collection of systems \cite{wang2018safe,cortez2019control,wu2022general} including automated vehicles \cite{ames2014control,seo2022safety}. 
%CBF takes positive values inside a set that we wish to remain in as a safety task, and it is zero on the boundary. 
CBF-based methods render a control system safe by imposing a condition on the control input
that ensures the non-negativeness of the CBF at all times, hence safety.
When this condition is used as a constraint to modify a nominal controller to the closest safe controller, we obtain the so-called \emph{safety filters}, which inspire our controller integration scheme.

\begin{figure*}[t]
  \centering
  \includegraphics[width=1\linewidth]{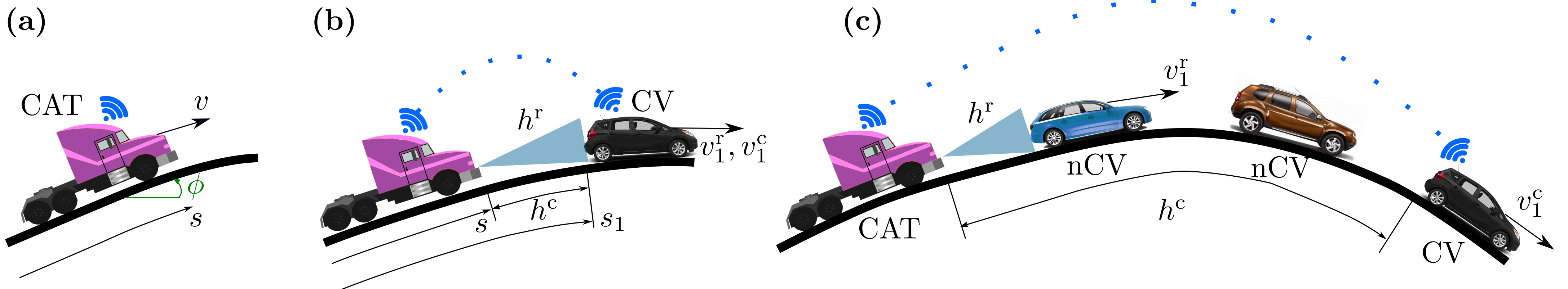}\\
  \caption{Connected automated truck (CAT) driving on a highway in three scenarios. (a) There is no preceding vehicle. (b) The closest preceding vehicle is a connected vehicle (CV) detected by the range sensor and connectivity simultaneously. (c) There are non-connected vehicles (nCVs) between the CAT and CV, and range sensor and connectivity detect different vehicles.}
  \label{fig:scheme}
  \vspace{-0 mm}
\end{figure*}

Our second contribution is validating the overall control design experimentally. We implement the resulting safe integration scheme on a full-scale connected automated truck (CAT).
To highlight the proposed scheme's flexibility, we employ two safety-oriented controllers: one responding to the closest vehicle with a range sensor and another interacting with a connected vehicle ahead . 
We consider energy efficiency as the performance metric to be optimized and provide the design details of a performance-based cruise controller tracking an optimal-in-energy speed profile.
We report experiments conducted in a fully controlled environment, i.e., a closed test track, and in a highway scenario open to interventions from other human-driven vehicles. 
The scheme proposed here serves as a framework to easily integrate sophisticated controllers that take advantage of connectivity to improve the desired aspects of CAV technology.

The organization of the paper is as follows. In Section~\ref{sec:systemdescription}, we give a detailed description of the driving scenario of a CAT, and we present models representing the system. 
In Section~\ref{sec:controller}, we first introduce the safe integration scheme for the general longitudinal control problem of CAVs. 
Then, we give the design steps of an energy optimal nominal controller and a safety-oriented controller for a CAT. 
Section~\ref{sec:NPG} gives our first on-track experimental results validating the proposed controller structures and the integration scheme.
Next, we present the experimental results on a public highway in Section~\ref{sec:Kentucky}.
Finally, we conclude the paper with conclusions in Section~\ref{sec:conclusion}.

%%%%%%%%%%%%%%%%%%%%%%%%%%%%%%%%%%%%%%%%%%%%%%%%%%%%%%%%%%%%%%%%%%%%%%%%%%%%%%%%%%%%%%%%%%%%%%%%%%%%%%%%%
\section{System Description and Modeling}
\label{sec:systemdescription}

Here we describe the driving scenario and the truck's sensor and control systems. Then we give dynamical models representing the system.

%%%%%%%%%%%%%%%%%%%%%%%%%%%%%%%%%%%%%%%%
\subsection{System Description}
\label{sec:scenario}

This study considers a scenario where a CAT drives on a highway  with changing elevation.
The CAT either drives with no influence from the preceding traffic, as shown in Fig.~\ref{fig:scheme}(a), or it follows a preceding vehicle, as in Fig.~\ref{fig:scheme}(b)-(c). 
Onboard the CAT, there is a range sensor for detecting preceding vehicles and a communication module that enables connectivity with other connected vehicles (CVs).
We consider scenarios with one CV ahead, but the proposed controller scheme can be extended to multiple CVs.
The vehicle that the CAT follows as the closest preceding vehicle may be the CV, as shown in Fig.~\ref{fig:scheme}(b), or there may be other non-connected vehicles (nCVs) in between the CAT and CV, as in Fig.~\ref{fig:scheme}(c).

\begin{figure*}[t]
  \centering
  \includegraphics[width=\linewidth]{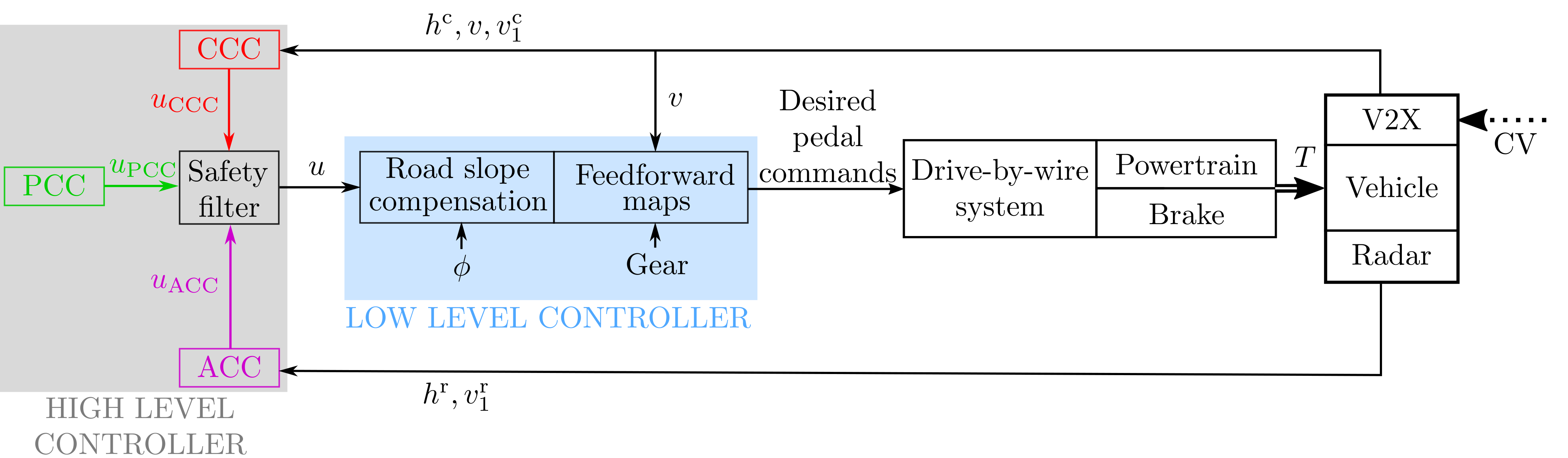}\\
  \caption{Block diagram representing the control architecture implemented on the connected automated truck.}
  \label{fig:blockdiagram}
  \vspace{-0 mm}
\end{figure*}

Our goal is to control the longitudinal motion of the CAT based on the information from the range sensor and communication module. We want to maintain safety as the primary concern while attaining the desired driving performance when possible. In this study, we consider energy efficiency as the single performance criterion, but we propose a general structure offering solutions for multiple performance objectives. 
To achieve our goal, we implement the control system illustrated in the block diagram of Fig.~\ref{fig:blockdiagram}. We retrofit the CAT with drive-by-wire actuators, which control the powertrain and brake subsystems based on the driver's pedal inputs \cite{Isermann__drivebywire:02}. 
We intervene the drive-by-wire system by replacing the driver's inputs with the \textit{desired pedal positions} calculated by a proposed longitudinal controller. 

The longitudinal controller we implement consists of a two-layer architecture with high and low levels. The high-level controller calculates the \textit{desired longitudinal acceleration}, denoted by $u$, based on a high-level controller goal. 
The high-level controller block in Fig.~\ref{fig:blockdiagram} highlights some examples of these controllers and a safe-integration scheme called a safety filter, which will be detailed in Section~\ref{sec:controller}.

\begin{figure}[b]
  \centering
  \includegraphics[width=1\linewidth]{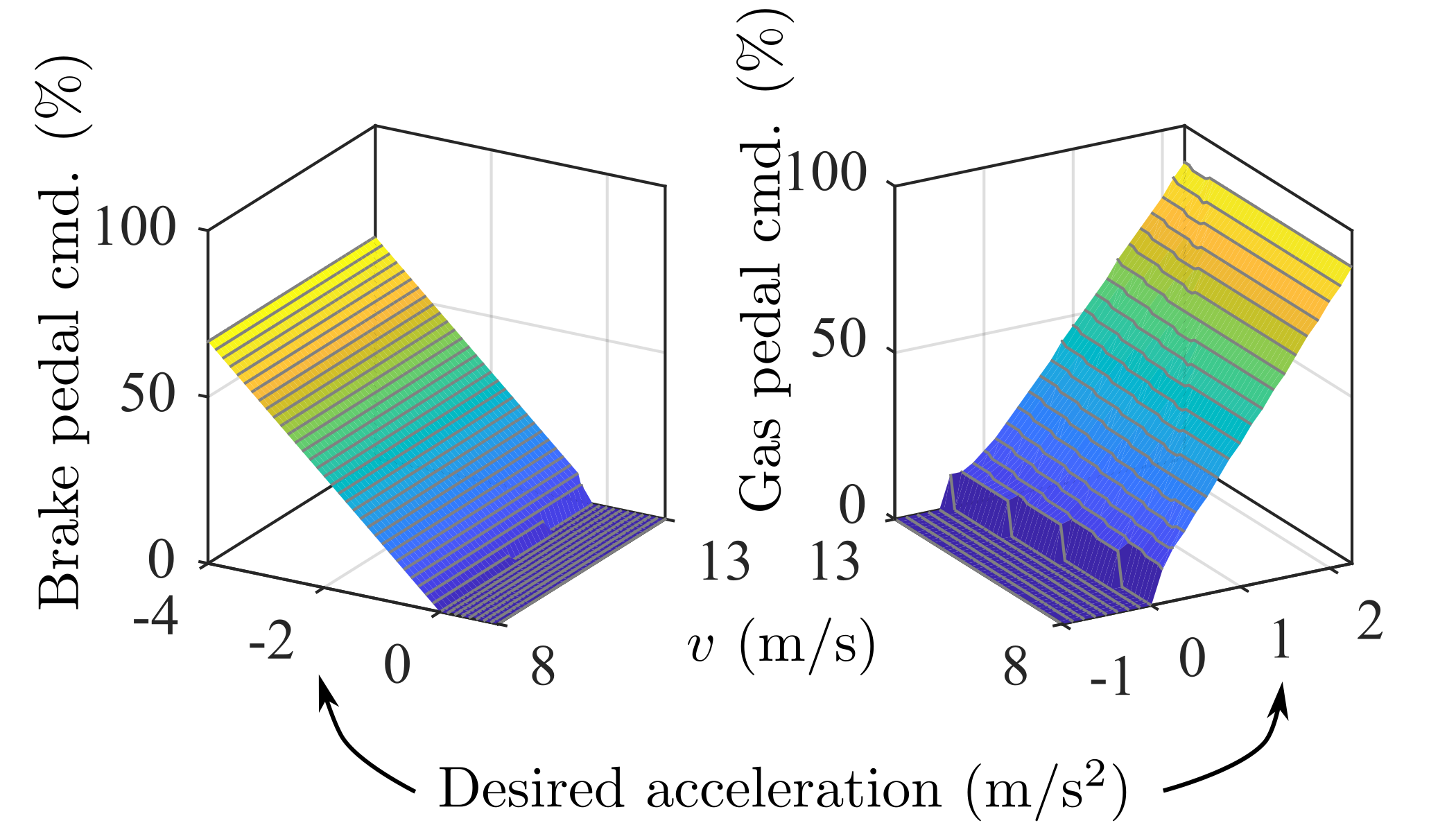}\\
  % \caption{Feedforward maps utilized by the low level controller in gear 7 to obtain pedal positions that track desired acceleration commands.}
  \caption{Feedforward maps utilized by the low-level controller outputing the pedal positions to track the desired acceleration commands. The right panel shows the map for gear 7.}
  \label{fig:FeedforwardMaps}
  \vspace{-0 mm}
\end{figure}

The low-level controller finds the corresponding desired pedal commands to track the desired acceleration commands as closely as possible.
Specifically, the low-level controller uses experimental data containing the acceleration response to pedal commands.
These data were generated at various speed levels, gears, and pedal commands in an open-loop fashion. 
The resulting relationships are inversely encoded in the low-level controller as feedforward maps which give the corresponding pedal commands to achieve the commanded acceleration based on speed and gear.
For example, Fig.~\ref{fig:FeedforwardMaps} depicts the feedforward maps utilized in gear 7. Note that the experimental data used in the feedforward map calculations do not capture the effect of the road slope. Thus, we compensate for the gravitational effect before applying the feedforward maps using road slope information obtained from elevation data. These data were collected via GPS along the particular road sections utilized prior to performing the experiments. 

%To calculate the high-level acceleration command, the control system relies on sensory feedback from the communication module and range sensor. 
The control system relies on sensory information from the communication module and range sensor to calculate the high-level acceleration command.
The communication module can measure GPS coordinates, and the ground speed $v$ of the CAT. 
It also receives information from the CV, including its ground speed $v_1^{\rm c}$ and GPS coordinates. 
These coordinates are used to calculate the longitudinal bumper-to-bumper headway $h^{\rm c}$ between two vehicles \cite{ge2018experimental}.
The range sensor is mounted on the front bumper and measures the headway $h^{\rm r}$ between the closest preceding vehicle and the CAT. 
It also measures the relative speed between vehicles, which is added to $v$ to get the ground speed $v_1^{\rm r}$ of the preceding vehicle. 
For simplicity, we will drop the superscripts `c' and `r' when there is no ambiguity and use them only to emphasize the difference between the source of the data (connectivity or range sensor) when needed.

%%%%%%%%%%%%%%%%%%%%%%%%%%%%%%%%%%%%%%%%
\subsection{Modeling}
\label{sec:models}

\begin{table}[b]
\centering
\begin{tabular}{|rl|rl|rl|}
\hline
\textbf{$R$}              & 0.5 m       &  
\textbf{$m$}              & 9000 kg     &
\textbf{$m_{\rm eff}$}    & 9157 kg         
\\ 
\hline
\textbf{$\gamma$}         & 0.006       & 
\textbf{$k_{\rm air}$}    & 3.84 kg/m   & 
\textbf{$g$}              & 9.81 m/s$^2$    
\\ 
\hline
\textbf{$\underline{u}$}  & 4 m/s$^2$   &
\textbf{$\overline{u}$}   & 2 m/s$^2$   & 
\textbf{$\overline{P}$}   & 93 kW       
\\ 
\hline
\end{tabular}
\caption{CAT parameters used in this study corresponding to a 2011 International ProStar+ Class-8 truck manufactured by Navistar Corporation \cite{navistar2021prostar}.}
\label{tab:truckparameters}
\vspace{-0 mm}
\end{table}

Here, we introduce system models utilized for control design. First principles are used to derive the longitudinal vehicle dynamics for a rear-axle-driven truck without headwind \cite{cakmakci__book:12}:
\begin{align}
\label{eq:vehiclemodel}
\begin{split}
    \dot{s} =&~ v, \\
    \dot{v} =&~ \dfrac{ T_{\rm w}}{R m_{\rm eff}} - \underbrace{\dfrac{ mg }{m_{\rm eff}} \big(\sin(\phi(s))+\gamma \cos(\phi(s))\big) }_{f_1(\phi(s))}  -  
    \underbrace{\dfrac{k_{\rm air}}{m_{\rm eff}}v^2}_{f_2(v)},
\end{split}
\end{align}
where $s$ is the distance traveled by the front bumper of the CAT along the road, $v$ and $\dot{v}$ are the longitudinal speed and acceleration, and $\phi(s)$ is the road slope changing along the road; see Fig.~\ref{fig:scheme}(a). Parameters in the model are the tire radius $R$, truck mass $m$, effective mass ${m_{\rm eff}=m+I/R^2}$ (incorporating the mass moment of inertia $I$ of rotating elements), rolling resistance $\gamma$, air drag coefficient $k_{\rm air}$, and gravitational acceleration $g$. For this study we use parameters corresponding to a truck without a trailer, given in Table~\ref{tab:truckparameters}. 

Here $T_{\rm w}$ denotes the net wheel torque applied at the  rear axle, consisting of a positive driving and a negative braking component. To work with these components using units of acceleration for convenience, we utilize the conversion ${T_{\rm w}= \hat{u} R m_{\rm eff}}$ and reformulate \eqref{eq:vehiclemodel} as
\begin{align}
\label{eq:vehiclemodel2}
\begin{split}
\dot{s} =&~ v,
\\
\dot{v} =&~ \hat{u} - f_1(\phi(s)) - f_2(v).
\end{split}
\end{align}
We remark that the scaled torque input $\hat{u}$ can be split as
\begin{equation}
\label{eq:u}
 \hat{u} = u_{\rm dr} + u_{\rm br}, 
\end{equation}
where ${u_{\rm dr}\geq0}$ and ${u_{\rm br}\leq0}$ represent the scaled driving and braking torques, respectively. 
This will be utilized when constructing the performance-based controller further below.
Assuming that the low-level controller compensates the functions $f_1$ and $f_2$, the model \eqref{eq:vehiclemodel2} reduces to
\begin{align}
\label{eq:vehiclemodel_control}
\begin{split}
    \dot{s} =&~ v, \\
    \dot{v} =&~ u,
\end{split}
\end{align}
where $u$ represents the desired acceleration prescribed by the high-level controller. 

We describe the motion of preceding vehicles using the kinematic model:
\begin{align}   
\label{eq:precedingvehicle}
\begin{split}
    \dot{s}_1 &= v_1, \\
    \dot{v}_1 &= a_1,
\end{split}
\end{align}
where $s_1$ is the distance traveled by the rear bumper of the preceding vehicle in consideration (see Fig.~\ref{fig:scheme}(b)-(c)), while $v_1$ and $a_1$ denote the corresponding speed and acceleration. To utilize the distance headway in the high-level controller we first define ${h \triangleq s_1 - s}$, and use \eqref{eq:vehiclemodel_control} and \eqref{eq:precedingvehicle} to obtain the car-following model:
\begin{align}   
\label{eq:carfollowingmodel}
\begin{split}
    \dot{h} &= v_1 - v, \\
    \dot{v} &= u, \\
    \dot{v}_1 &= a_1.
\end{split}
\end{align}
We note that the whether the preceding vehicle in consideration
is a CV or an nCV we may add the superscripts `c' and `r' for $h$ and $v_1$ to highlight the feedback's source; see Fig.~\ref{fig:scheme}(b)-(c).

In order to take into account the physical limitations in the powertrain and brake we limit the desired acceleration using
\begin{align}
\label{eq:adeslimits}
    -\underline{u} \leq u \leq \min \left\{ \overline{u} , \frac{\overline{P}}{m_{\rm eff}\,v} \right\}.
\end{align}
Here $\underline{u}$ is the maximum deceleration limit corresponding to the maximum brake torque. The term $\overline{u}$ denotes the maximum acceleration corresponding to the maximum torque applied on the driven wheels. The desired acceleration is also limited by the maximum power $\overline{P}$ of the powertrain. 
We list the values of $\underline{u}$, $\overline{u}$, and $\overline{P}$ in Table~\ref{tab:truckparameters}.
Finally, we also consider 
\begin{equation}
    \label{eq:vlimits}
    \underline{v} \leq v \leq \overline{v},
\end{equation}
where the speed limits $\underline{v}$ and $\overline{v}$ are determined based on the road curvature, surface conditions, and legal limitations.

%%%%%%%%%%%%%%%%%%%%%%%%%%%%%%%%%%%%%%%%%%%%%%%%%%%%%%%%%%%%%%%%%%%%%%%%%%%%%%%%%%%%%%%%%%%%%%%%%%%%%%%%%
\section{Controller Design}
\label{sec:controller}

In this section we formally define safety for a connected automated vehicle in the single-lane scenario. 
We introduce the notion of a safety filter based on the safety-critical control task. Then, we show the design steps for a safety-oriented connected cruise controller and an optimal-in-energy nominal controller for a connected automated truck.

%%%%%%%%%%%%%%%%%%%%%%%%%%%%%%%%%%%%%%%%
\subsection{Safe Controller Integration Scheme}
\label{sec:safeintegration}

The safety task for a CAV within a single lane (i.e., no overtaking considered) is to follow the preceding vehicle while maintaining \textit{at least} a certain critical distance at all times. This can be formulated as
\begin{equation}  
\label{eq:safetytask}
   h(t) \geq \rho \left(v(t),v_1(t) \right), \quad \forall t \geq 0,
\end{equation}
where the function $\rho$ gives the critical distance based on the speeds of the CAV and the preceding vehicle. 

There are many possible approaches to specify the critical distance $\rho$, such as the minimum time-to-collision \cite{Li__TTC:20} or minimum time headway \cite{Ayres__timeheadway:01}. Furthermore, one may take the input capabilities of the CAV, given by \eqref{eq:adeslimits}, into consideration to provide feasibility to the resulting safety-critical controller as described in \cite{Chaozhe:18}, which is then extended to a smooth quadratic function in \cite{alan2022control}. 
The scheme proposed in this study can be applied to a large class of safety tasks under Assumption~\ref{ass:rho}.

\begin{assumption}    
\label{ass:rho}
The critical distance the CAV shall keep from the preceding vehicle strictly increases with the speed of the ego vehicle; that is, for a continuously differentiable $\rho$ we have
\begin{equation}     
    \derp{}{v}\rho(v,v_1)>0,
\end{equation}
for all ${v\in[\underline{v},\overline{v}]}$ and ${v_1\geq0}$.
\end{assumption}
The reasoning behind this assumption is that the faster the vehicle travels, the larger distance it shall keep.

Safety tasks in the form of \eqref{eq:safetytask} are often studied in the context of \textit{set invariance} in the literature \cite{blanchini2008set}, where the state of a dynamical system should remain inside a prescribed set for all time. To be specific, consider a set given as 
\begin{equation}
\label{eq:safeset}
    \C \triangleq \{ [h,v,v_1]^\top \in\R^3 ~|~ h-\rho(v,v_1)\geq0 \},
\end{equation}
for the car-following setup \eqref{eq:carfollowingmodel}.
For an initial condition ${[h(0),v(0),v_1(0)]^\top\in\C}$, if ${[h(t),v(t),v_1(t)]^\top\in\C}$ for all ${t\geq 0}$, then we say the system is \textit{safe} with respect to the set ${\C}$, which ensures  safety defined as \eqref{eq:safetytask}. 

{Control barrier functions (CBFs)} offer a solution to synthesize controllers for problems of this type \cite{ames2017control}. A detailed description is given in Appendix~\ref{app:CBFandSafetyFilter}. In simple terms, a CBF renders a system safe by providing a condition for the controller to satisfy (cf.~\eqref{eq:app_Kcbf}). When this condition is used as a constraint to modify a nominal control input to the closest safe input (cf.~\eqref{eq:app_QP}), the resulting controller is called \textit{safety filter}. 
Safety filters operate instantaneously (without any horizon) and offer easy-to-implement solutions, especially for single input systems such as the longitudinal control of a CAV. 

We use the model \eqref{eq:carfollowingmodel} for a general CAV, which, along with Assumption~\ref{ass:rho}, yields the CBF-based safety filter
\begin{equation}   
\label{eq:safeintegration}
    u = \min \left\{ u_{\rm safe}, u_{\rm nom} \right\},
\end{equation}
see  Appendix~\ref{app:CBFandSafetyFilter} for calculation details.
This algorithm will be used as the proposed {\em safe controller integration scheme}.
Here $u_{\rm nom}$ is a nominal controller that can be tailored to optimize an aspect of the system without considering safety, such as optimal-in-energy driving. The term $u_{\rm safe}$ denotes a safety-critical controller. The safety-critical controller can be any algorithm certified to keep the system safe by satisfying the controller constraint provided by the CBF. 
When the nominal controller is smaller than the safety-critical controller, it can be shown to satisfy the safety task \eqref{eq:safetytask}. Therefore, the safety filter passes the nominal controller without modification to maintain performance. However, the safety filter switches to the safety-critical controller when it becomes smaller than the nominal one to ensure that safety is not violated. 

In the remainder of this section, we present details about implementing the safe controller integration scheme \eqref{eq:safeintegration} to the longitudinal control problem of a connected automated truck with a safety-critical controller and an energy-efficient nominal controller.
We note that solving the control problem of modifying a nominal controller to the nearest safety-critical controller provides us with a form of $u_{\rm safe}$ (cf.~\eqref{eq:app_usafe}) for a specific selection of $\rho$ used in the safety task \eqref{eq:safetytask}. 
However, since we do not specify $\rho$, we choose to replace the CBF-based safety-critical controller with another safety-oriented controller shown to be safe experimentally under various working conditions.

%%%%%%%%%%%%%%%%%%%%%%%%%%%%%%%%%%%%%%%%%
\subsection{Safety-oriented Connected Cruise Controller}

A connected cruise controller structure is utilized as the safety-oriented controller for the safety filter \eqref{eq:safeintegration}. 
This type of controller was studied extensively for stability and string stability under time delays and system uncertainties \cite{Linjun2016CCCtimedelay,Wubing2017CCCstability,Hajdu2020CCCrobust}, and it was shown to be safe experimentally for different driver behaviors \cite{ge2018experimental}. 

\begin{figure}[t]
  \centering
  \includegraphics[width=\linewidth]{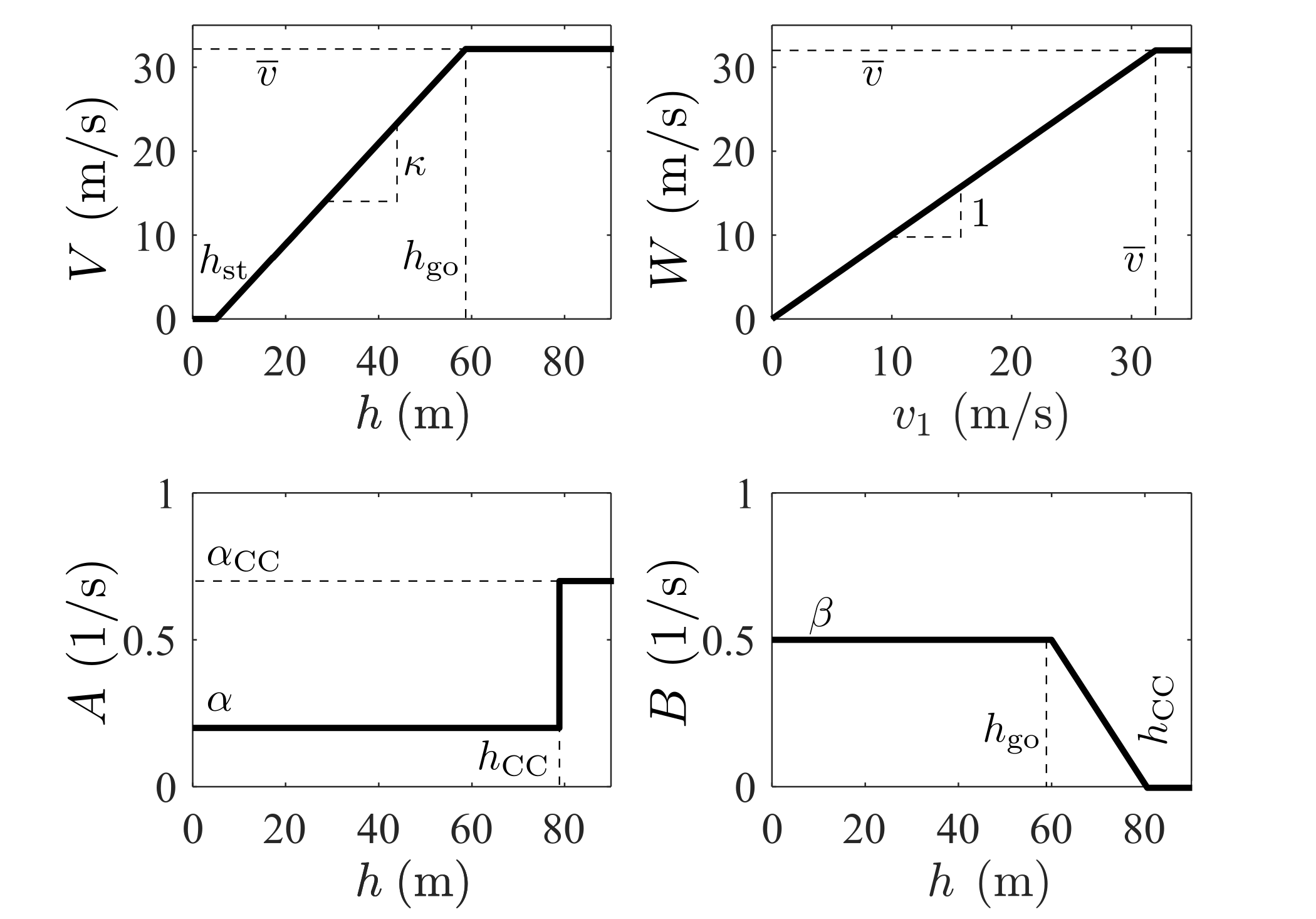}\\
  \caption{Range policy \eqref{eq:rangepolicy}, speed policy \eqref{eq:speedpolicy}, controller gain functions \eqref{eq:A(s,h1)} and \eqref{eq:B(s,h1)} with parameters used in Section~\ref{sec:Kentucky}. }
  \label{fig:controllerFigures}
  \vspace{-0 mm}
\end{figure}

The controller structure is given as 
\begin{equation}
    \label{eq:safecontroller}
    u_{\rm safe}(h,v,v_1) = A(h) ( V(h) - v ) + B(h) ( W(v_1)-v ),
\end{equation}
where the first term gives the desired acceleration based on the speed error associated with the \textit{range policy} 
\begin{equation}
\label{eq:rangepolicy}
    V(h) = 
    \begin{cases}
    0                           &~{\rm{if}} \quad h \leq h_{\rm st}, \\
    \kappa\,(h-h_{\rm st})    &~{\rm{if}} \quad h_{ \rm st} < h < h_{\rm go}, \\
    \overline{v}             &~{\rm{if}} \quad h \geq h_{\rm go},
    \end{cases}
\end{equation}
%for all ${s\geq0}$, 
see Fig.~\ref{fig:controllerFigures}(a).
Parameter $h_{\rm st}$ denotes the stopping distance, and $\kappa$ is the gradient determining the relationship between the distance headway and the target speed. The value ${h_{\rm go} \triangleq h_{\rm st} + \overline{v}/\kappa}$ is the distance, after which the range policy outputs the maximum speed $\overline{v}$ as the target speed.
The second term in \eqref{eq:safecontroller} yields the desired acceleration based on the relative speed subject to the \textit{speed policy} 
\begin{equation}
\label{eq:speedpolicy}
    W(v_1)  =  \min \left\{ v_1,\overline{v} \right\},
    % \begin{cases}
    % v_1                 &~{\rm{if}} \quad v_1 \leq \overline{v}, \\
    % \overline{v}        &~{\rm{if}} \quad v_1 > \overline{v},
    % \end{cases}
\end{equation}
which is introduced to put a bound on the speed error in case the preceding vehicle moves faster than the speed limit, see Fig.~\ref{fig:controllerFigures}(b). 

Terms $A$ and $B$ determine the gains associated with the speed errors based on the range policy and the relative speed. We choose 
\begin{equation}
\label{eq:A(s,h1)}
    A(h) =  
    \begin{cases}
    \alpha           &~{\rm{if}} \quad h \leq h_{\rm CC},\\
    \alpha_{\rm CC}  &~{\rm{if}} \quad h > h_{\rm CC},
    \end{cases}
\end{equation}
\begin{equation}
\label{eq:B(s,h1)}
    B(h)  = 
    \begin{cases}
    \beta                           &~{\rm{if}} ~ h \leq h_{\rm go}, \\
    \beta \dfrac{h_{\rm CC}-h}{h_{\rm CC}-h_{\rm go}} &~{\rm{if}} ~ h_{\rm go} < h < h_{\rm CC}, \\
    0                               &~{\rm{if}} ~ h \geq h_{\rm CC},
    \end{cases}
\end{equation}
as depicted in Fig.~\ref{fig:controllerFigures}(c)-(d).
Notice that the effect of the preceding vehicle's speed gradually diminishes for larger headways. For ${h \geq h_{\rm CC}}$
we have ${A(h)=\alpha_{\rm CC}}$ and ${B(h)=0}$, that is, \eqref{eq:safecontroller} yields the constant speed cruise controller $\alpha_{\rm CC}(\overline{v}-v)$ tracking the maximum speed $\overline{v}$ with a constant gain $\alpha_{\rm CC}$. We will specify the distance ${\delta \triangleq h_{\rm CC}-h_{\rm go}}$ in the experiments discussed further below.

We remark that given a specific selection of a differentiable function $\rho$, one may show that the controller \eqref{eq:safecontroller} certifies the safety of the set $\C$  defined in \eqref{eq:safeset} by satisfying \eqref{eq:app_Kcbf} for a large enough $h_{\rm st}$ and small enough $\kappa$.

%%%%%%%%%%%%%%%%%%%%%%%%%%%%%%%%%%%%%%%%%
\subsection{Performance-based Nominal Controller}
\label{sec:nominalcontroller}

In this study, we consider the `wheels-to-distance' energy efficiency as the performance aspect of a truck \cite{sciarretta2015optimal} and employ a \textit{predictive cruise controller (PCC)} \cite{He__fuel:16}. PCC minimizes the mechanical energy input by considering the constraints on powertrain output and speed \eqref{eq:adeslimits},\eqref{eq:vlimits} and taking advantage of the variable road slope over a preview distance ${s\in[0,\,s_{\rm f}]}$, where $s_{\rm f}$ denotes the end of the horizon. 

In PCC, we formulate an optimal control framework to find the  optimal drive and brake components $u_{\rm dr}^*$ and $u_{\rm br}^*$ in \eqref{eq:u} while minimizing the mechanical energy. The vehicle model \eqref{eq:vehiclemodel2} is used as a constraint in the optimal control framework.
Since the road slope depends on the position, i.e., $\phi(s)$,
the implementation variables are converted from the time domain to spatial domain using ${\der{t}{s}=\frac{1}{v}}$ while assuming positive speed: 
\begin{equation}\label{eq:speedlimit}
0 < \underline{v}  \leq v \leq  \overline{v},
\end{equation}
cf.~\eqref{eq:vlimits}. These lead to the model
\begin{equation}
\label{eq:PCCoptimization_vehicle}
    \der{v}{s}=\frac{u_{\rm dr}+u_{\rm br}}{v} - \frac{f_1(\phi(s)) + f_2(v)}{v}.
\end{equation}
Similar to \eqref{eq:adeslimits} we consider the following limitations on decision variables:
\begin{align}
     \label{eq:acclimit}
    0 \leq  u_{\rm dr} &\leq \min\left\{ \overline{u}_{\rm dr},  \frac{\overline{P}}{m_{\rm eff}v}\right\}, 
    \\
     \label{eq:breaklimit}
    -\underline{u}_{\rm br} \leq  u_{\rm br} &\leq 0, 
    \\
    \label{eq:PCCoptimization_udrubr}
    u_{\rm dr} u_{\rm br} & = 0,
\end{align}
where $\underline{u}_{\rm br}$ and $\overline{u}_{\rm dr}$ correspond to the maximum driving and braking torques, and $\overline{P}$ denotes the maximum power of the powertrain. The constraint \eqref{eq:PCCoptimization_udrubr} is introduced to ensure that throttle and brake are not active simultaneously.
The cost function we wish to minimize is selected as the mechanical energy input per unit effective mass:
\begin{equation}
\label{eq:energy}
    w(t) = \int_0^{t} u_{\rm dr} v {\rm d}\tilde{t},
\end{equation}
where the integrand refers to the power input per unit effective mass. Note that we do not include the brake torque component in \eqref{eq:energy} to prevent possible negative consumption when brakes are active. 

Converting the energy integral to the spatial domain \eqref{eq:speedlimit}-\eqref{eq:energy} result in the optimal control problem \cite{He__fuel:16}:
\begin{align}
\label{eq:PCCoptimization}
    ( u^*_{\rm dr}, u^*_{\rm br}) = & \argmin_{( u_{\rm dr}, u_{\rm br})}  \int_0^{s_{\rm f}}  u_{\rm dr} {\rm d}\tilde{s},  
    \\
    \text{ subject to } ~ 
    & \der{v}{s}=\frac{ u_{\rm dr}+ u_{\rm br}}{v} - \frac{f_1(\phi(s)) + f_2(v)}{v}, \notag 
    \\
    & 0 \leq  u_{\rm dr} \leq \min\left\{ \overline{u}_{\rm dr},  \frac{\overline{P}}{m_{\rm eff}v}\right\}, \notag 
    \\
    & -\underline{u}_{\rm br} \leq   u_{\rm br} \leq 0, \notag 
    \\
    &  u_{\rm dr}  u_{\rm br} = 0, \notag \\
    & \underline{v} \leq v \leq \overline{v}, \notag \\
    & v(0) = v_{0}, \notag \\
    & v(s_{\rm f}) = v_{\rm f}, \notag \\
    & \int_{0}^{s_{\rm f}} \frac{1}{v(s)} \,{\rm d}s \leq \overline{t}_{\rm f}, \notag
\end{align}
with boundary conditions ${v(0)=v_{0}}$ and ${v(s_{\rm f})=v_{\rm f}}$ for speed. The last constraint with maximum travel time $\overline{t}_{\rm f}$ is introduced to ensure that the travel time is not sacrificed for better energy efficiency. We typically obtain boundary conditions and the maximum travel time from a benchmark run driven by an expert human driver. 

To solve the optimization problem \eqref{eq:PCCoptimization}, 
we rely on road slope data obtained as follows.
We use GPS measurements of the benchmark run to calculate the discretized travel distance values $s_i$ along the road. The slope ${\phi(s_i)=\sin^{-1}\left(\frac{{\rm d} E(s_i)}{{\rm d} s}\right)}$ is calculated from the elevation  $E(s_i)$ measured at corresponding points $s_i$ via numerical differentiation. Finally, the open-source interior point solver IPOPT \cite{IPOPT2006} is used to solve the resulting nonlinear programming problem offline, yielding the optimal inputs $u_{\rm dr}^*(s_i)$ and $u_{\rm br}^*(s_i)$ as well as the optimal speed profile $v_{\rm PCC}^*(s_i)$.
Rather than directly implementing the optimal inputs, we employ a feedback controller strategy to reject potential disturbances emerging from the inaccuracies in the low-level controller. 
Thus, a variable-speed cruise controller with a constant gain $\alpha_{\rm CC}$ is implemented:
\begin{equation}
    \label{eq:nominalcontroller}
    u_{\rm PCC}(s,v) = \alpha_{\rm CC} \big(v_{\rm PCC}(s) - v\big),
\end{equation}
where $v_{\rm PCC}(s)$ is calculated from $v_{\rm PCC}^*(s_i)$ via interpolation for any given ${s\in[0,\,s_{\rm f}]}$.

Energy efficiency evaluation for different experimental runs is carried out by calculating the cost function \eqref{eq:energy} along the road. Since we do not implement $ u_{\rm dr}^*$ directly, we need to calculate the $ u_{\rm dr}$ values corresponding to the implemented controller effort. We calculate $ u_{\rm dr}$ using the vehicle dynamics \eqref{eq:vehiclemodel2} and the measured speed, acceleration, and road slope data:
\begin{equation}
\label{eq:drivetorque_from_accel}
     u_{\rm dr} = \max \big\{ 0, \dot{v} + f_1( \phi(s)) + f_2(v) \big\},
\end{equation}
where ${\max\{0,\cdot\}}$ is introduced to ensure ${u_{\rm dr}\geq0}$.

%%%%%%%%%%%%%%%%%%%%%%%%%%%%%%%%%%%%%%%%%
\subsection{Implemented Safe Controller Integration Scheme for CAT}

Having introduced controller strategies \eqref{eq:safecontroller} and \eqref{eq:nominalcontroller} focusing on different tasks, we now integrate them using the safety filter concept \eqref{eq:safeintegration}.
Utilizing the structure \eqref{eq:safecontroller}, we employ two safety-oriented controllers distinguished by their feedback source. We name the controller 
${u_{\rm safe}(h^{\rm r},v,v_1^{\rm r})}$,
which relies on the range sensor data, as the adaptive cruise controller (ACC) and denote it by ${u_{\rm ACC}}$. The controller ${u_{\rm safe}(h^{\rm c},v,v_1^{\rm c})}$,
which employs the connectivity-based data, is called the connected cruise controller (CCC) and is denoted by ${u_{\rm CCC}}$. The nominal controller is selected as PCC \eqref{eq:nominalcontroller}, yielding:
\begin{equation}
\label{eq:safetyfilter}
    \begin{split}
    u(s,h^{\rm c},h^{\rm r},v,v_1^{\rm c},v_1^{\rm r}) = 
    \min \{ u_{\rm ACC}&(h^{\rm r},v,v_1^{\rm r}), \\
    u_{\rm CCC}&(h^{\rm c},v,v_1^{\rm c}), \\
    u_{\rm PCC}&(s,v)  \},
    \end{split}
\end{equation}   
see Fig.~\ref{fig:blockdiagram}. 
In \eqref{eq:safetyfilter}, PCC is utilized as long as it is considered safe, and switched to either of the safety-oriented controllers occurs based on their feedback. 

It is noted that the CV detected through connectivity may or may not be the closest preceding vehicle in open traffic, as shown in Fig.~\ref{fig:scheme}(b)-(c).
If the CV is the closest preceding vehicle, ACC and CCC respond to the same vehicle (with slight differences based on sensor readings). When other non-connected vehicles are between the CAT and CV, the range sensor detects the closest preceding vehicle, and ACC responds accordingly while CCC follows the CV.
We remark that the beyond-line-of-sight detection capabilities of the connectivity forebode significant improvement in both safety and energy efficiency \cite{vahidi2018energy,Minghao2022CCC,Wong2021trafficforecase}. In this study, we focus on proving the efficacy of the proposed safe controller integration method. Thus we leave the work of utilizing more sophisticated connectivity-based controller structures as future work.

%%%%%%%%%%%%%%%%%%%%%%%%%%%%%%%%%%%%%%%%%%%%%%%%%%%%%%%%%%%%%%%%%%%%%%%%%%%%%%%%%%%%%%%%%%%%%%%%%%%%%%%%%
\section{On-Track Experiments}
\label{sec:NPG}

In this section, we describe the experimental results obtained on a closed track to validate the proposed controller structures. 
In these experiments, the CAT was controlled to follow an optimal-in-energy speed profile without a preceding vehicle, or to follow a preceding CV according to the scenarios depicted in Fig.~\ref{fig:scheme}(a) and (b).
After describing the experimental setup and procedure, results for the nominal controller, the safety-oriented controller, and the integrated controller with the proposed safety filter scheme are introduced.

%%%%%%%%%%%%%%%%%%%%%%%%%%%%%%%%%%%%%%%%%
\subsection{Details about Experimental Setup and Procedure}
\label{sec:experimentalsetup}

\begin{figure}[t]
  \centering
  \includegraphics[width=1\linewidth]{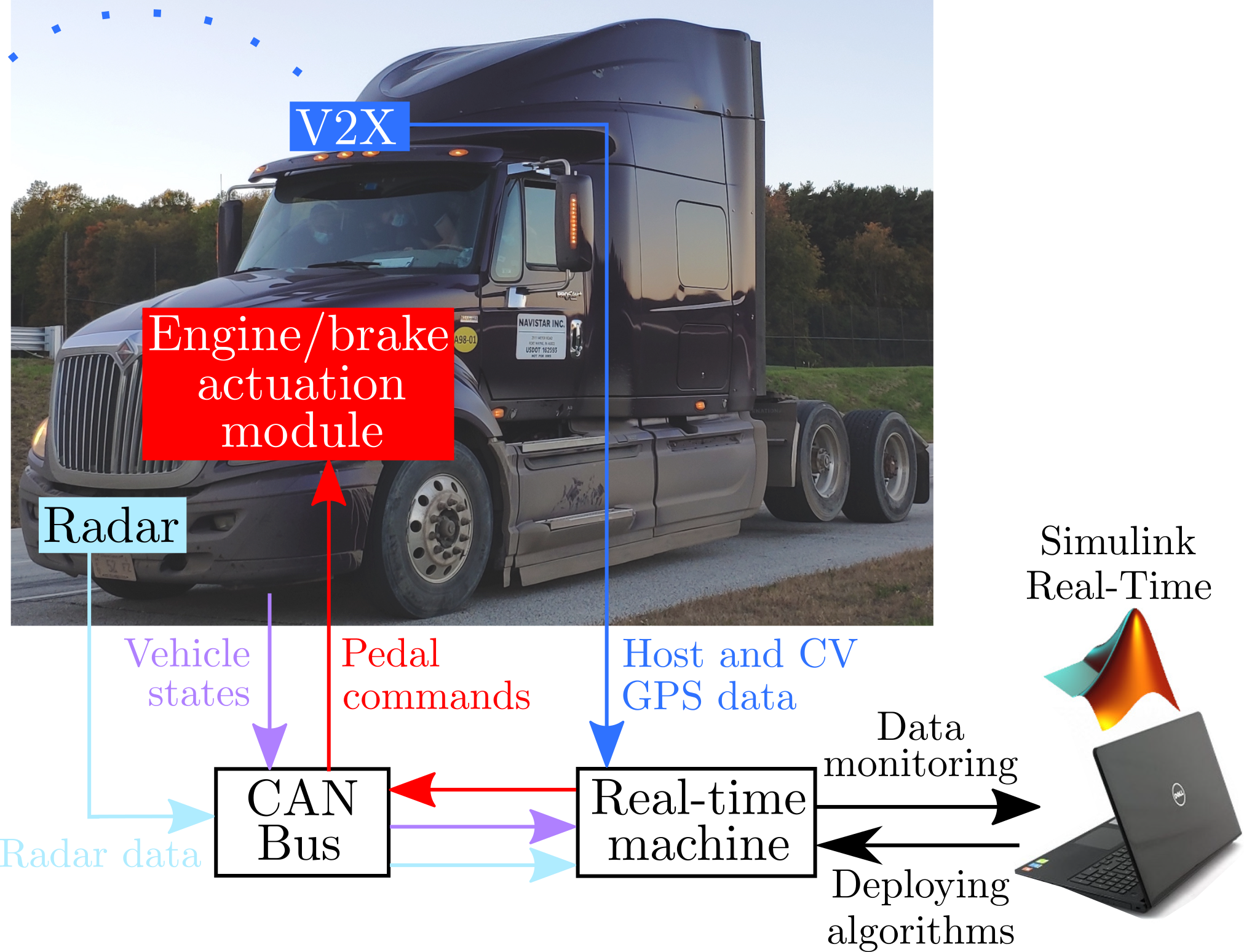}\\
  \caption{The connected automated truck (CAT) is used for experiments and the information flow between its units.}
  \label{fig:Hardware}
  \vspace{-0 mm}
\end{figure}

We used a 2011 International ProStar+ Class-8 truck developed by Navistar \cite{navistar2021prostar} as the CAT; see Fig.~\ref{fig:Hardware}. The truck has engine and brake actuation modules configured to follow the desired pedal commands sent through the vehicle-CAN bus following J1939 CAN protocol \cite{J1939/71_202002}. The vehicle states, such as the wheel speed, gear position, engine torque and rpm, and brake pressure are also available on the CAN bus under the same protocol. The truck is equipped with a Mobile Real-Time Targeting Machine developed by Speedgoat \cite{speedgoat2021mrtu}, which reads the vehicle states from the vehicle-CAN bus, runs the control algorithms in Simulink Real-Time, computes the corresponding desired pedal commands, and sends them to the vehicle-CAN bus. A personal computer was connected to the real-time machine deploying algorithms in one direction and monitoring data in another. The computer could also abort the mission at any time, giving the control for pedals back to a human driver. Steering was carried out by an expert human driver all the time.

\begin{figure}[t]
  \centering
  \includegraphics[width=1\linewidth]{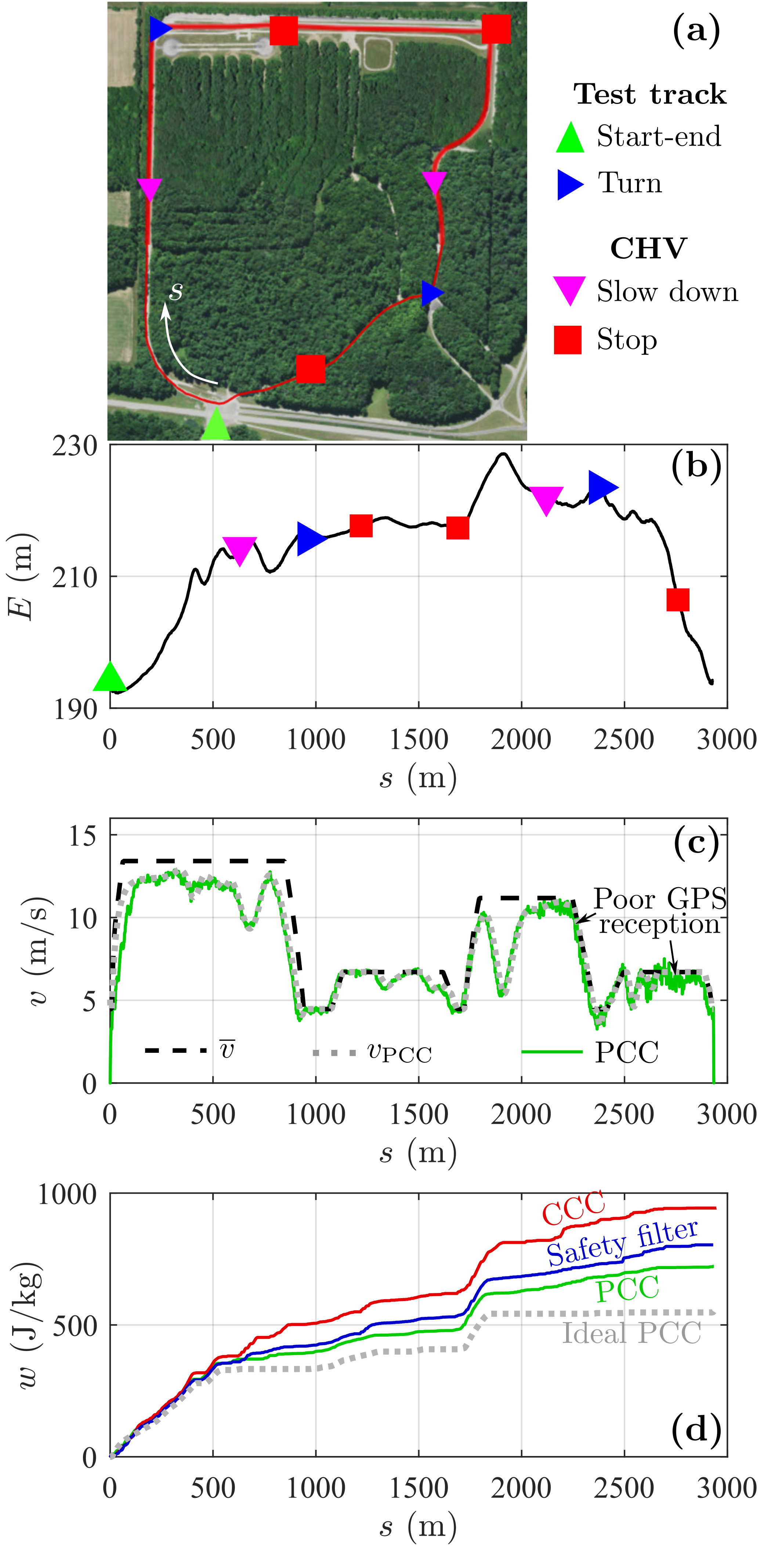}
  \caption{(a) Test track used for experiments with landmarks indicating details about the CV speed profile design. (b) Elevation profile $E$ with corresponding landmark positions. (c) Speed limit $\overline{v}$, optimal-in-energy speed profile $v_{\rm PCC}$, and measured speed profile $v$ in the PCC experiments. (d) Energy consumption curves $w$ calculated from experimental data using \eqref{eq:energy} for different controllers. }
  \label{fig:NPG_PCC}
  \vspace{-0 mm}
\end{figure}

A radar was utilized as the range sensor. The radar sent messages containing the headway and relative speed information belonging to the closest preceding vehicle to the vehicle-CAN bus.
Both the truck and the CV were equipped with a Vehicle-to-Everything (V2X) communication onboard unit (OBU) developed by Commsignia \cite{commsignia2021v2x}. These units provide GPS coordinates and GPS-based speed measurements and support peer-to-peer communication between vehicle. The V2X OBU on the truck received data packets broadcasted by CV's V2X OBU at a rate of 10 Hz, containing GPS coordinates and speed information of the CV. The real-time communicated between the CAT's V2X OBU and the Speedgoat was established using user datagram protocol (UDP).

The on-track experiments were conducted at Navistar Proving Grounds, a test track closed to the public in New Carlisle, Indiana, USA.
The route used is shown in Fig.~\ref{fig:NPG_PCC}(a) with the GPS trace of a benchmark run (red loop in the clockwise direction). The start-end point of the experiments is marked by a green triangle. Using GPS-based elevation measurements from multiple benchmark runs, an average elevation profile of the track was estimated as shown in Fig.~\ref{fig:NPG_PCC}(b). 
Blue triangles on the map and the elevation plot indicate sharp turns; these were considered when designing the speed limit $\overline{v}$ along the road, see the black dashed line in Fig.~\ref{fig:NPG_PCC}(c). We also used a smaller speed limit in the middle section due to poor road surface quality. 

The loop of nearly 3000 meters of length was discretized using the GPS trace of a benchmark run, resulting in GPS data points approximately 2.5 meters away from each other. The optimal control problem \eqref{eq:PCCoptimization} was solved offline, and the corresponding optimal-in-energy speed profile $v_{\rm PCC}$ is plotted in Fig.~\ref{fig:NPG_PCC}(c) as a gray dotted curve. 
The resulting energy consumption per unit effective mass \eqref{eq:energy}, i.e., cost function in \eqref{eq:PCCoptimization},  is depicted in Fig.~\ref{fig:NPG_PCC}(d) as a gray dotted curve. Notice that the optimal profile requires very little energy after 1800 m and utilizes the gravitational potential energy to finish the drive while obeying the speed limit. We used parameters given in Table~\ref{tab:NPG_parameters} for all the on-track experiments detailed in this section.

%%%%%%%%%%%%%%%%%%%%%%%%%%%%%%%%%%%%%%%%%
\subsection{Results} 
\label{sec:experimentdescription}

\begin{table}[t]
\centering
\begin{tabular}{|rl|rl|rl|}
\hline
\textbf{$h_{\rm st}$}      & 5 m       & 
\textbf{$\kappa$}          & 0.6 1/s   &    
\textbf{$\delta$}          &  20 m   
\\ 
\hline
\textbf{$\alpha$}          &  0.4 1/s  &
\textbf{$\beta$}           & 0.5 1/s   &  
\textbf{$\alpha_{\rm CC}$} &  0.9 1/s     

\\ 
\hline
\textbf{$\overline{u}_{\rm dr}$}    & 2 m/s$^2$ & 
\textbf{$\underline{u}_{\rm br}$}   & 4 m/s$^2$ &  \textbf{$\underline{v}$}   &  2.5 m/s        
\\ 
\hline
\end{tabular}
\caption{Controller parameters used for on-track experiments. }
\label{tab:NPG_parameters}
\vspace{-0 mm}
\end{table}

First, we implemented the PCC controller \eqref{eq:nominalcontroller} on the CAT with the computed optimal speed profile and
without any preceding vehicle on the track. The
results are depicted in Fig.~\ref{fig:NPG_PCC}(c) as a green curve. One may observe good speed-tracking performance in the cruise control, which verifies the performance of the low-level controller. The corresponding energy consumption per unit mass calculated from  \eqref{eq:energy} is shown in Fig.~\ref{fig:NPG_PCC}(d) as a green curve. One may notice the difference between the ideal energy consumption (gray dotted curve) and the experimental one (green curve). This gap is partly due to the powertrain dynamics omitted in the optimal control problem \eqref{eq:PCCoptimization}, and partly due to the noise in the experimental data, especially towards the end of the run. This noise was due to poor GPS reception, caused by a dense canopy, and it was amplified by the numerical differentiation employed to obtain the acceleration in \eqref{eq:drivetorque_from_accel}. When integrated per \eqref{eq:energy}, the noise results in a positive drift due to the function $\max \{0,\cdot \}$. Since we employ a comparative analysis among different controller runs in this study, we ignore these imperfections and focus on how well a controller performs in terms of energy consumption compared to other controllers in the experiments.

\begin{figure}[t]
  \centering
  \includegraphics[width=1\linewidth]{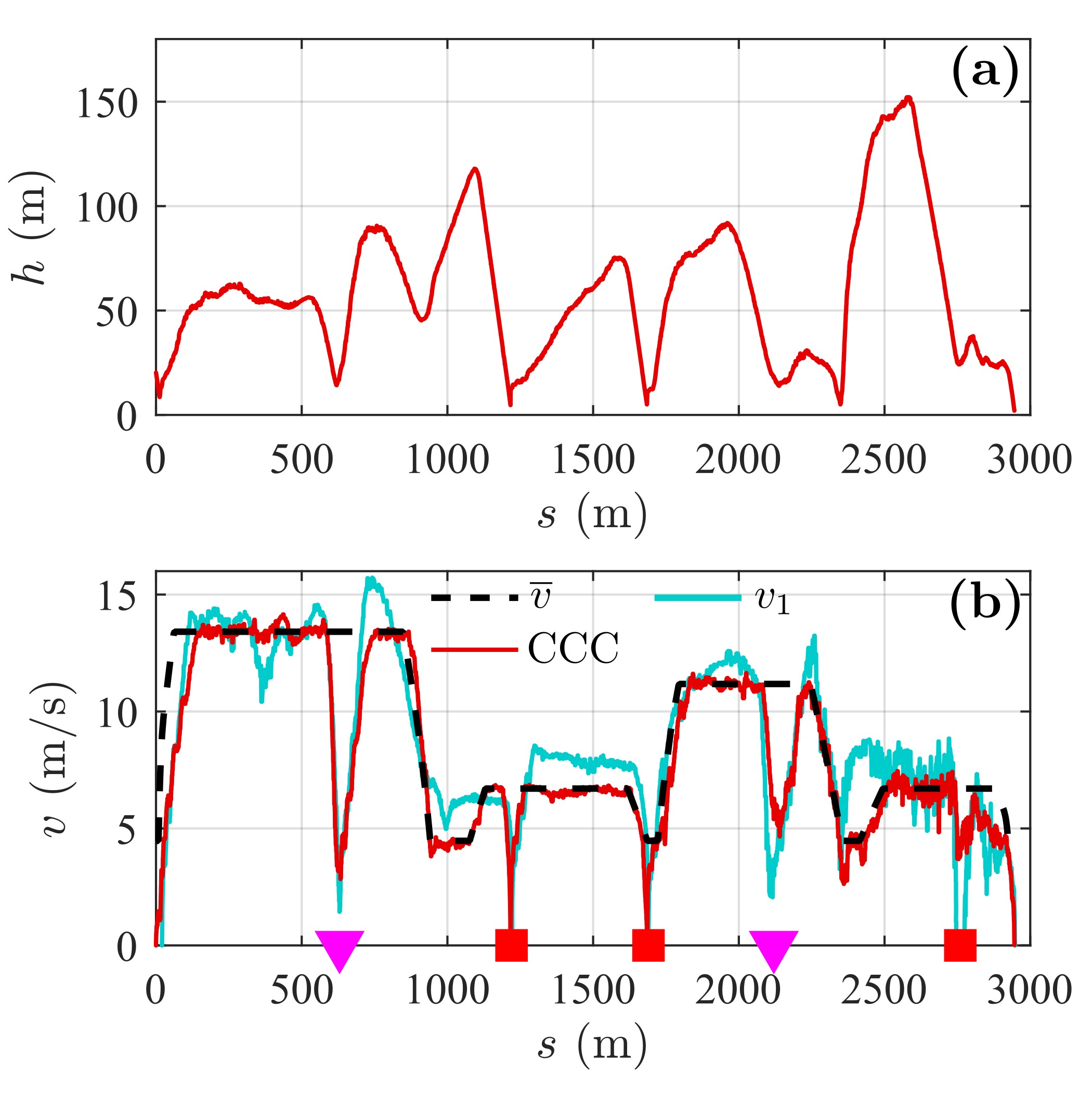}
  \caption{Experimental results from a CCC run. (a) Measured distance headway $h$. (b) Speed limit $\overline{v}$, CV speed profile $v_1$, and measured CAT speed profile $v$ in the experiment.}
  \label{fig:NPG_ACC}
  \vspace{-0 mm}
\end{figure}

Next, we experimentally validated the safety-oriented controller \eqref{eq:safecontroller} such that the CAT traveled behind a CV. 
Since the test track was closed to public traffic and no non-connected vehicles were present, we only implemented CCC utilizing connectivity-based data. 
This setup, which is equivalent to the radar-based ACC in this scenario as depicted in Fig.~\ref{fig:scheme}(b), allowed us to record the GPS trace and speed of the CV in a particular run and later re-play this record for the CAT. 
These experiments not only prevented a physical collision in the case of a malfunction in hardware or software but also provided us with 
consistent preceding vehicle motion across
different runs so that the repeatability could be evaluated solely on the merits of controllers. In particular, we designed a speed profile $v_1$ for the preceding vehicle that imitates a heavy traffic scenario with multiple slowdowns (marked by magenta triangles) and stops(marked by red squares). Fig.~\ref{fig:NPG_ACC}(b) shows the resulting CV speed profile as a cyan curve. 

\begin{figure}[t]
  \centering
  \includegraphics[width=1\linewidth]{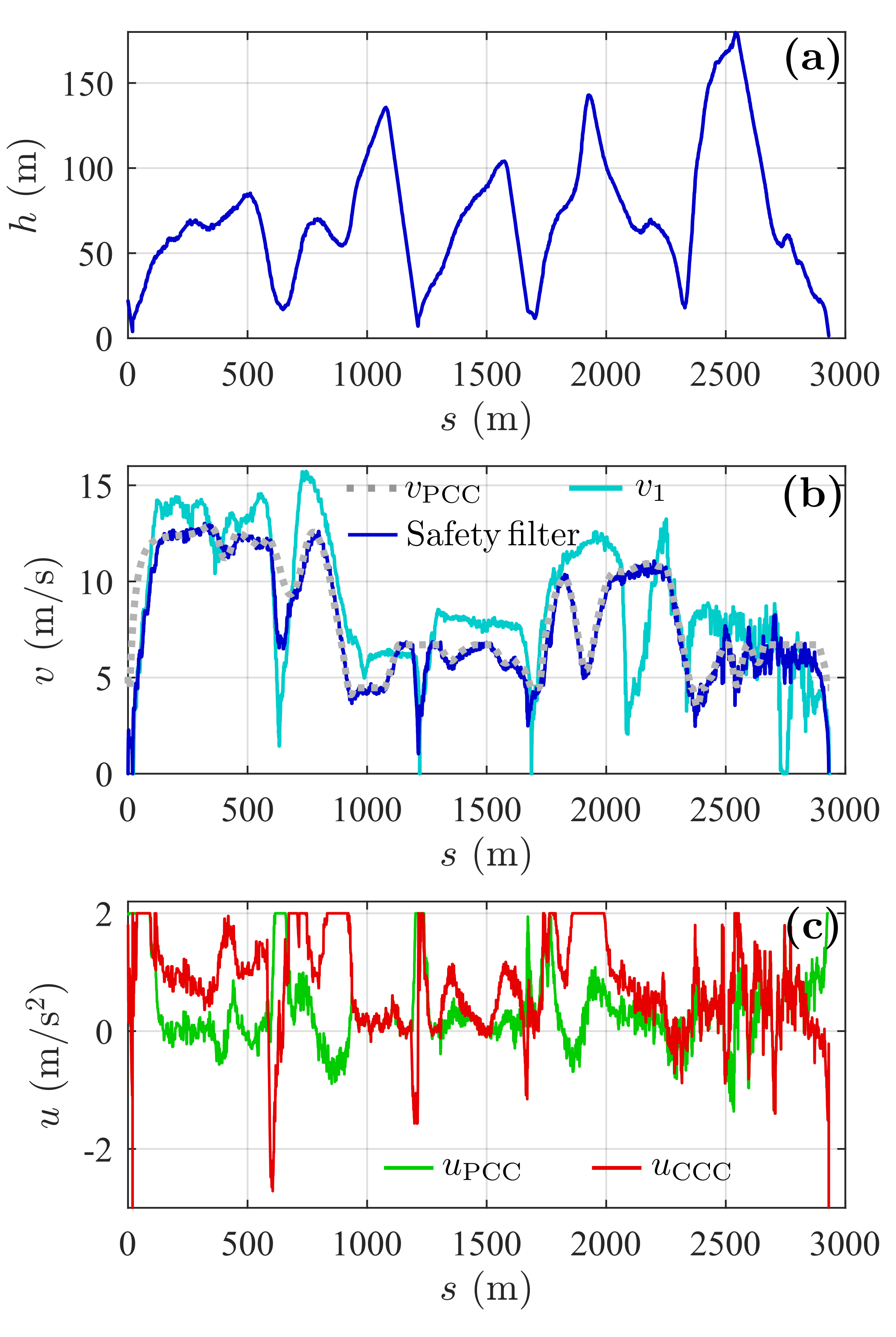}
  \caption{Experimental results while applying the proposed safe controller integration scheme. (a) Measured distance headway $h$. (b) CV speed profile $v_1$, optimal-in-energy speed profile $v_{\rm PCC}$, and meaured CAT speed profile $v$ in the experiment. (d) Controller outputs $u_{\rm PCC}$ and $u_{\rm CCC}$. The safety filter passed the minimum of these.}
  \label{fig:NPG_INT}
  \vspace{-0 mm}
\end{figure}

Experimental results for a CCC run are given in Fig.~\ref{fig:NPG_ACC}, where panel (a) shows the measured distance headway and panel (b) depicts the measured speed, both as red curves. 
CCC keeps the system safe by maintaining a positive headway throughout the run. Moreover, the truck successfully obeys the speed limit even when the CV moves faster than the limit. The energy consumption throughout the run is depicted in Fig.~\ref{fig:NPG_PCC}(d) as a red curve, where the car-following is observed to yield more significant energy input rates than PCC around 700 meters and around 1800 meters due to the significant braking and acceleration triggered by the CV's motion.
Even though a slightly better performance in energy efficiency could be attained through carefully designing CCC parameters \cite{he2019fuel}, it would continue to 
be suboptimal in energy efficiency. 

Finally, we implemented the safety filter \eqref{eq:safetyfilter} utilizing PCC and CCC as the nominal and safety-oriented controllers (without the radar-based ACC). Results are given in Fig.~\ref{fig:NPG_INT}, where panel (a) shows the measured distance headway and panel (b) depicts the measured speed, both as blue curves. Panel (c) displays controllers $u_{\rm PCC}$ and $u_{\rm CCC}$ as green and red curves, respectively. Similar to the previously presented CCC result, a positive headway was maintained throughout the run thanks to the safety filter switching to CCC in critical moments around 600 m, 1200 m, and 1650 m. 
In these moments, the acceleration command of CCC becomes smaller than PCC, responding to the other vehicle and ensuring safety. At other times PCC was active since it suggested more energy-efficient driving rather than following the preceding vehicle. This switch yielded a 15\%  energy consumption savings compared to the CCC run, cf. Fig.~\ref{fig:NPG_PCC}(d). 

We repeated these experiments 8-10 times for each configuration with the same conditions (the same CV speed profile was played back). Headway never became negative in any of the CCC and safety filter runs. The results in energy efficiency are summarized in Table~\ref{tab:NPG_results} under the label `Recorded' in terms of energy savings compared to the CCC runs. All runs are consistent with our initial findings, with only slight deviations among different runs. 

Afterward, similar experiments were conducted with a physical human-driven preceding vehicle reenacting the CV speed profile $v_1$ with close accuracy. Similarly, the distance headway never reached zero in any of the CCC and safety filter runs. Moreover, similar energy efficiency results were obtained with a slight increase in the standard deviations due to the slight increase in deviations between the motion profiles of the CV; see results given in Table~\ref{tab:NPG_results} under the label `Live'. A video illustratively summarizing on-track experiments is available online \cite{SupplementVideo}.

\begin{table}[t]
\centering {
\begin{tabular}{|c|c|c|}
\hline
\textbf{CV data}             & \textbf{Controller}  & \begin{tabular}{@{}c@{}}\textbf{Energy saving} \\ \textbf{compared to CCC}\end{tabular} \\ \hline
\multirow{2}{*}{Recorded} & PCC                  & 23\%  $\pm$  4\%   \\ \cline{2-3} 
                          & Safety Filter        & 18\%  $\pm$  3\%   \\ \cline{1-3}
\multirow{2}{*}{Live}     & PCC                  & 25\%  $\pm$  4\%   \\ \cline{2-3} 
                          & Safety Filter        & 18\%  $\pm$  5\%   \\ \hline
\end{tabular}
}
\caption{Summary of energy efficiency in multiple experiments conducted using a recorded CV speed profile (cyan curve in Fig.~\ref{fig:NPG_ACC}(b)) and a physical (live) CV reenacting the same profile.}
\label{tab:NPG_results}
\end{table}

%%%%%%%%%%%%%%%%%%%%%%%%%%%%%%%%%%%%%%%%%%%%%%%%%%%%%%%%%%%%%%%%%%%%%%%%%%%%%%%%%%%%%%%%%%%%%%%%%%%%%%%%%
\section{Highway Experiments}
\label{sec:Kentucky}

Having proved the efficacy of the safety filter in a closed test track, we proceed to validate the proposed structure on a public highway.

%%%%%%%%%%%%%%%%%%%%%%%%%%%%%%%%%%%%%%%%%
\subsection{Details about the Experimental Procedure}

In these experiments, the CAT was driven on a public road amongst non-connected human-driven vehicles (nCVs), and one connected vehicle (CV) was driven by an expert driver from our team. Thus, scenarios in both Fig.~\ref{fig:scheme}(b) and (c) occurred. For the CAT, we used the same experimental setup as described in the previous section; see Fig.~\ref{fig:Hardware}. Note that in these experiments, the CAT was able to utilize the information from both the radar and connectivity.

\begin{figure}[!t]
  \centering
  \includegraphics[width=1\linewidth]{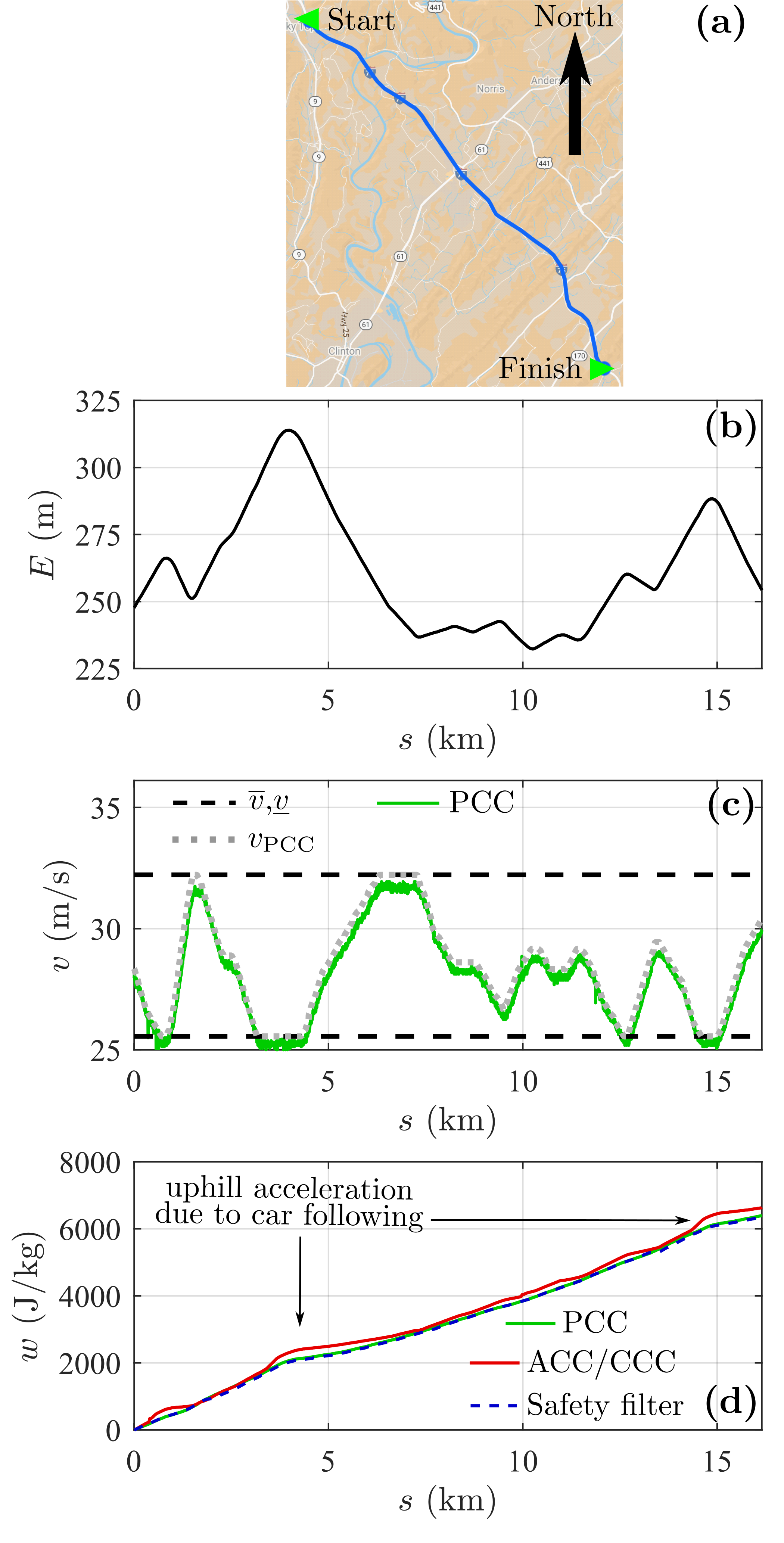}
  \caption{(a) Section of I-75 used for the highway experiments. (b) Elevation profile $E$. (c) Speed limits $\underline{v}$ and  $\overline{v}$ within which the PCC box operates, optimal-in-energy speed profile $v_{\rm PCC}$, and measured CAT speed profile $v$ in the PCC experiments. (d) Energy consumption curves $w$ calculated from experimental data using \eqref{eq:energy} for different controllers. }
  \label{fig:Kentucky_PCC}
  \vspace{-5 mm}
\end{figure}

A section on the Interstate 75 highway was selected for highway experiments as shown in Fig.~\ref{fig:Kentucky_PCC}(a). We had previously collected elevation data on this road using four GPS sensors attached to four vehicles. The averaged elevation profile is shown in Fig.~\ref{fig:Kentucky_PCC}(b) with two significant hills enabling the optimal-in-energy controller framework to leverage gravitational potential energy. 

Note that in highway experiments, we employed a PCC box, which is hardware that hosts the commercialized version of the PCC algorithm described in Section \ref{sec:nominalcontroller} and is available for trucks manufactured by Navistar \cite{pccbox_navistar}, as a part of our nominal controller.
While details about the algorithm running under the PCC box are omitted here due to its confidential nature, it utilizes the road slope information to calculate the optimal-in-energy speed profile similar to the optimal control framework \eqref{eq:PCCoptimization}, but in a rolling horizon fashion. 
We used the desired speed values calculated by the PCC box as $v_{\rm PCC}$ in our variable-speed cruise controller structure \eqref{eq:nominalcontroller}. The PCC box enables us to choose the speed limits $\overline{v}$ and $\underline{v}$ for our system; please refer to Table~\ref{tab:Kentucky_parameters} for these and all the other parameters used in highway experiments. 

\begin{table}[!b]
\centering
\begin{tabular}{|rl|rl|}
\hline
\textbf{$h_{\rm st}$}     & 5 m       & 
\textbf{$\alpha_{\rm CC}$}& 0.7 1/s
\\ 
\hline
\textbf{$\kappa$}         & 0.8 1/s   & 
\textbf{$\underline{v}$}  & 25 m/s 
\\ 
\hline
\textbf{$\delta$}         & 20 m      & 
\textbf{$\overline{v}$}   & 32 m/s  
\\ 
\hline
\textbf{$\alpha$}         & 0.2 1/s   & 
\textbf{$\overline{u}$}   & 2 m/s$^2$   
\\ 
\hline
\textbf{$\beta$}          & 0.5 1/s   & 
\textbf{$\underline{u}$}  & 3 m/s$^2$   
\\ 
\hline
\end{tabular}
\caption{Controller parameters used for the highway experiments. }
\label{tab:Kentucky_parameters}
\vspace{-0 mm}
\end{table}

The resulting optimal-in-energy speed profile is shown in Fig.~\ref{fig:Kentucky_PCC}(c) as a gray dotted curve. Similar to the on-track experiments, one may notice the variation of the optimal speed responding to the elevation changes along the road while obeying to the speed limits.

%%%%%%%%%%%%%%%%%%%%%%%%%%%%%%%%%%%%%%%%%
\subsection{Results}

First, we implemented the PCC \eqref{eq:nominalcontroller} using the speed profile $v_{\rm PCC}$ attained from the PCC box. Results are depicted in Fig.~\ref{fig:Kentucky_PCC}(c) as a green curve. Observe the good speed-tracking performance with a small steady-state error (up to 0.5 m/s). This error arose due to aerodynamics, which was more prominent at higher speeds than the low-speed experiments conducted when constructing the feedforward maps in the low-level controller. 
%A higher controller gain $\alpha_{\rm CC}$ could be picked to reduce this tracking error further.
The energy consumption profile corresponding to the PCC run is displayed in Fig.~\ref{fig:Kentucky_PCC}(d) as a green curve; here, favorable energy consumption can be observed even at the at uphill regions thanks to the optimal speed profile taking into account the elevation profile of the hilly terrain. Table~\ref{tab:Kentucky_results} summarizes the final energy consumption values in the highway experiments.

\begin{table}[b]
  \centering {
    \begin{tabular}{|c|c|c|c|}
    \hline
    \textbf{Controller}     & \begin{tabular}{@{}c@{}}\textbf{Final energy} \\ \textbf{consumption} \\ \textbf{value} \end{tabular}       & \begin{tabular}{@{}c@{}}\textbf{Energy saving} \\ \textbf{compared to} \\ \textbf{ACC/CCC}\end{tabular}                 & \begin{tabular}{@{}c@{}}\textbf{Finish} \\ \textbf{time}\end{tabular}     \\
    \hline
    PCC                     & 6396 J/kg       & 3.6 \%     & 579 s              \\
    \hline
    ACC/CCC                 & 6635 J/kg       & -        & 577 s              \\
    \hline
    Safety filter           & 6350 J/kg       & 4.3 \%   & 583 s              \\
    \hline
    \end{tabular}%
    }
   \caption{Summary of highway experimental results.}
   \label{tab:Kentucky_results}
\end{table}

\begin{figure}[t]
  \centering
  \includegraphics[width=1\linewidth]{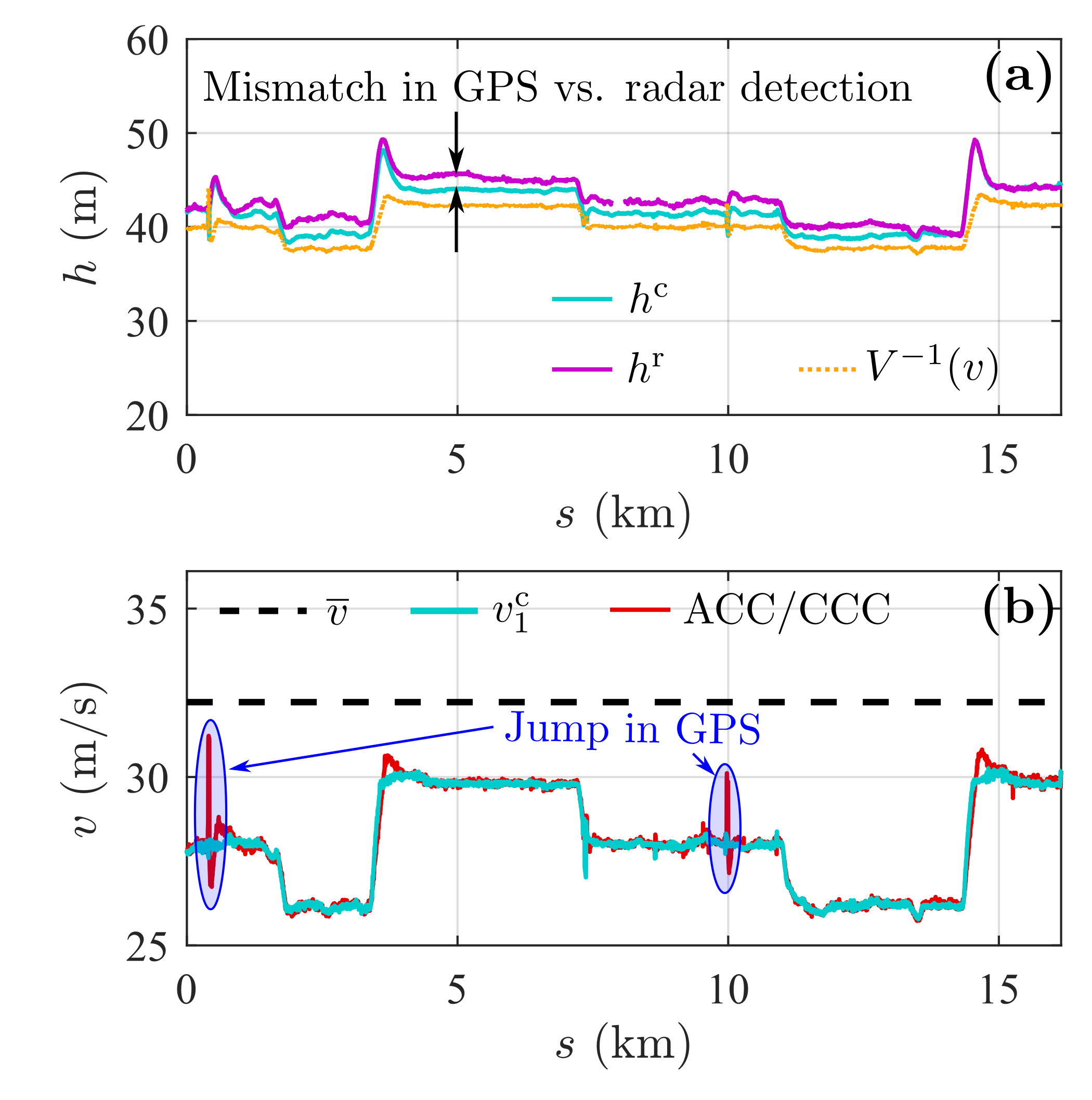}
  \caption{Highway experimental results for the ACC/CCC experiments. (a) Connectivity-based distance headway $h^{\rm c}$ and radar-based distance headway $h^{\rm r}$. (b) Speed limit $\overline{v}$, CV speed profile $v_1^{c}$ and CAT speed profile $v$ measured in the experiments.}
  \label{fig:Kentucky_CCC}
  \vspace{-0 mm}
\end{figure}

Next, we present the results of the highway experiment when employing only safety-oriented controllers: the radar-based ACC and connectivity-based CCC without the PCC (from here on referred to as ACC/CCC). Note that in this case we still apply the safety filter \eqref{eq:safetyfilter} to integrate ACC and CCC in a seamless fashion; see Fig.~\ref{fig:blockdiagram}. Similar to the on-track experiments, we employed a single CV agent in the scenario of Fig.~\ref{fig:scheme}(b) with a speed profile that includes uphill acceleration actions at around 3 km and 14.5 km; see Fig.~\ref{fig:Kentucky_CCC}(b). Another criterion in designing the CV speed profile was to have a similar finish time as the PCC run to avoid penalizing time for energy efficiency. 

The distance headways and the speeds for the ACC/CCC run are given in Fig.~\eqref{fig:Kentucky_CCC}(a) and (b), respectively. 
Good car-following performance can be observed in keeping the desired distance headway $V^{-1}(v)$ specified by the inverse of the range policy \eqref{eq:rangepolicy} (orange dotted curve in panel (a)) and in maintaining close track of the velocity of the preceding vehicle $v_1^{\rm c}$ (cyan curve in panel (b)). 
In panel (a) the headway measured by the radar $h^{\rm r}$ (magenta) and calculated from connectivity $h^{\rm c}$ (cyan) show a small mismatch (up to 3.5 m). 
The safety filter \eqref{eq:safetyfilter} handles these inconsistencies by passing the controller that demands the most safety-critical acceleration. 
We note the jumps in the measured data in two separate time instances, highlighted in Fig.~\ref{fig:Kentucky_CCC}(b). 
These jumps were merely a malfunction in GPS sensing, and they did not influence the experiments significantly. 

The energy consumption profile calculated for the ACC/CCC run given in Fig.~\ref{fig:Kentucky_PCC}(d) as a red curve, where the effect of the uphill acceleration due to following the energy-adverse CV speed profile is highlighted. The goal of finishing the course in a similar time as the PCC run was achieved as shown in Table~\ref{tab:Kentucky_results}, and the truck's similar initial and final speeds for both experiments imply that an energy saving of 3.6\% has been accomplished via using PCC over ACC/CCC.

\begin{figure}[t]
  \centering
  \includegraphics[width=1\linewidth]{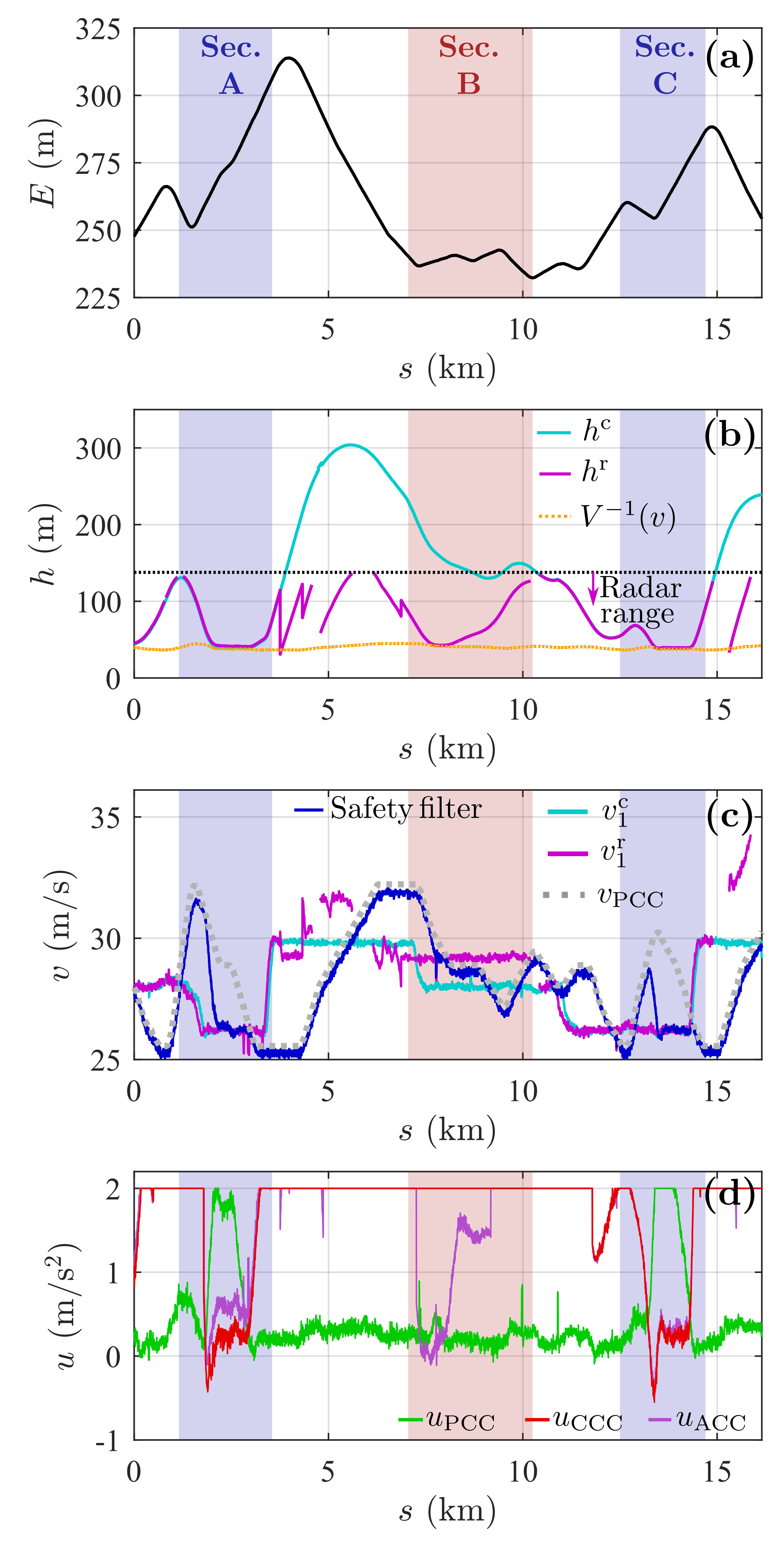}
  \caption{Highway experimental results with the safety filter \eqref{eq:safetyfilter} employed. (a) Connectivity-based distance headway $h^{\rm c}$ and radar-based distance headway $h^{\rm r}$. (b) CV speed based on GPS $v_1^{\rm c}$, preceding vehicle speed detected by radar $v_1^{\rm r}$, optimal-in-energy speed profile $v_{\rm PCC}$ and measured speed $v$ of the truck (blue). (d) Controller outputs $u_{\rm PCC}$, $u_{\rm CCC}$ and $u_{\rm ACC}$. The safety filter passed the minimum of these.}
  \label{fig:Kentucky_INT}
  \vspace{-0 mm}
\end{figure}

Finally, we implemented the safety filter \eqref{eq:safetyfilter} with the PCC, ACC, and CCC. The CV followed the same speed profile as above (cyan curve in Fig.~\ref{fig:Kentucky_CCC}(b)). Results are presented in Fig.~\ref{fig:Kentucky_INT}, where panel (a) shows the elevation profile of the road, while panel (b) depicts headway measured by the radar $h^{\rm r}$ (magenta) and calculated from connectivity $h^{\rm c}$ (cyan) as well as the target headway specified by the inverse of the range policy (orange dotted). 
When the cyan and magenta curves coincide, it indicates no vehicle between the truck and the CV; cf.~the scenario depicted in Fig.~\ref{fig:scheme}(b).
When engaged with a preceding vehicle, the safety filter successfully keeps the distance headway around the target value.
While the radar sensing range restriction can be seen in this plot, the communication between V2X units continued over a distance of 300 m.

Fig.~\ref{fig:Kentucky_CCC}(c) shows the speed signals of interest: the CV speed captured by connectivity $v_1^{\rm c}$ (cyan), the speed of the closest preceding vehicle detected by radar $v_1^{\rm r}$ (magenta), the optimal speed profile calculated by PCC box $v_{\rm PCC}$ (gray dotted) and the measured speed $v$ of the truck (blue). Panel (d) presents the controller outputs $u_{\rm PCC}$ (green), $u_{\rm CCC}$ (red), and $u_{\rm ACC}$ (magenta). 
Gaps in radar signals correspond to no vehicle detection in front, and sudden jumps indicate cut-ins from the other lanes. At some of these cut-ins, the lane-changing vehicle traveled faster than the truck (e.g., between 4-7 km and at 15.5 km), which was not considered safety-critical by ACC, and therefore the safety filter kept using the PCC. 
As a matter of fact, one may notice in panel (d) that PCC was the active controller throughout the majority of the run, resulting in a comparable energy consumption to the PCC run, as shown in Fig.~\ref{fig:Kentucky_PCC}(d) (blue dashed curve). The energy saving compared to the CCC run is 4.3\% without significantly increasing the course finish time, cf.~Table~\ref{tab:Kentucky_results}.

Switch to the safety-oriented controllers (i.e., to ACC or CCC) occurred at three separate locations: at 2 km, 7.5 km, and 13 km. While the first and the last engagements were between the CAT and the CV, in case of the middle one the truck responded to another non-connected vehicle traveling in the traffic.
In the first occurrence, highlighted as Sec. A in Fig.~\ref{fig:Kentucky_INT}, the PCC was initially the active controller. Then, the CV reduced its speed allowing the truck to catch up and engage. Consequently, the truck finished tracking the optimal speed profile and started following the CV with CCC (as the connectivity-based headway was reading a slightly smaller value than the radar-based headway, i.e., ${h^{\rm c}<h^{\rm r}}$). Then, the optimal speed profile gradually declined in the uphill climbing section, favoring the PCC over CCC in the safety filter and avoiding an uphill acceleration once the CV increased its speed.
In the second switch sequence, labeled as Sec. B, the ACC briefly engaged with a non-connected vehicle traveling between the CV and truck, see the scenario depicted in Fig.~\ref{fig:scheme}(c). With time, the optimal speed became less than the preceding vehicle's speed, resulting in the PCC becoming the active controller.
The last switching sequence at 13 km, highlighted as Sec. C, occurred in a similar order as Sec. A. A video highlighting the events occurring in Sec. A is available online \cite{SupplementVideo}.

%%%%%%%%%%%%%%%%%%%%%%%%%%%%%%%%%%%%%%%%%%%%%%%%%%%%%%%%%%%%%%%%%%%%%%%%%%%%%%%%%%%%%%%%%%%%%%%%%%%%%%%%%
\section{Conclusion}
\label{sec:conclusion}

In this paper, we proposed a simple yet effective scheme to integrate performance-based nominal controllers with safety-oriented controllers for the longitudinal control of connected automated vehicles. 
We implemented the proposed scheme on a real connected automated truck. We presented the design steps of a predictive cruise controller as a nominal controller and a connected cruise controller as a safety-oriented controller.
Then, we validated the proposed integration structure in two different experimental settings.
First, we used a closed test track to validate controllers separately and integrated them with the proposed integration scheme.
In these experiments we showed that the energy efficiency acquired through the predictive cruise controller can be integrated with the safety provided by the connected cruise controller. 
Finally, we verified these initial findings experimentally on a public highway. 
We showed that the safe controller integration scheme is also able handle interactions with other vehicles in traffic  while acquiring energy-efficient, yet safe driving on hilly terrain.

%%%%%%%%%%%%%%%%%%%%%%%%%%%%%%%%%%%%%%%%%%%%%%%%%%%%%%%%%%%%%%%%%%%%%%%%%%%%%%%%%%%%%%%%%%%%%%%%%%%%%%%%%
\appendices

\section{Control Barrier Functions and Safety Filter}  
\label{app:CBFandSafetyFilter}

Consider a nonlinear system of the form:
\begin{equation}
    \label{eq:app_system}
    \dot{x} = f(x)+g(x)\,u,
\end{equation}
with state ${x\in\R^n}$ and input ${u\in\R^m}$ along with a set ${\C \subset \R^n}$ defined as the 0-superlevel set of a continuously differentiable function ${b : \R^n \to \R}$:
\begin{align}
    \label{eq:app_C}
    \C \triangleq \left\{x \in \R^n ~\left|~ b(x) \geq 0 \right. \right\}.
\end{align}
System \eqref{eq:app_system} is said to be \textit{safe} with respect to the set $\C$ if the following holds: ${x(t_0) \in \C \implies x(t) \in \C}$ for all ${t\geq t_0}$. We name the set $\C$ as \textit{safe set}.

\begin{definition}[\textit{Control Barrier Function}, \cite{ames2017control}]
The function $b$ is a \textit{Control Barrier Function (CBF)} for \eqref{eq:app_system} on $\C$ if there exists ${\alpha\in\K}$ such that for all $x\in\R^n$:
\begin{equation}  
\label{eq:app_CBF}
     \sup_{u\in\R^m}  \underbrace{\left[ \nabla b(x)\cdot(f(x)+g(x)u) \right]}_{\dot{b}(x,u)}  > -\alpha(b(x)).
\end{equation}
\end{definition}
\noindent We note that a continuous function $\alpha$ is said to belong to \textit{class $\K$} (${\alpha\in\K}$) if ${\alpha(0)=0}$ and $\alpha$ is strictly monotonically increasing. As stated by Corollary~2 in \cite{ames2017control}, a controller from the set of controllers given as:
\begin{equation}
\label{eq:app_Kcbf}
    K_{\rm CBF}(x) \triangleq \left\{u \in \R^m ~\left|~\dot{b}(x,u) \geq -\alpha(b(x)) \right. \right\}
\end{equation}
renders the system \eqref{eq:app_system} safe with respect to $\C$. 

Consider the Quadratic Program (QP):
\begin{align}
\label{eq:app_QP}
\begin{split}
    u^*(x) = \,\,\underset{u \in \R^m}{\argmin}  &  \quad \frac{1}{2} \| u - u_{\rm nom}(x) \|_2^2  \\
    \mathrm{ s.t.} \quad & \quad \dot{b}(x,u) \geq -\alpha(b(x)),
\end{split}
\end{align}
which yields the notion of \textit{safety filters}. Here the controller $u_{\rm nom}$ denotes a nominal controller that has been designed to ensure performance without considering safety.
The safety filter outputs the nominal controller as long as it satisfies the CBF condition, i.e., ${u^*(x)=u_{\rm nom}(x)}$ for all $x$ such that ${u_{\rm nom}(x) \in K_{\rm CBF}(x)}$ holds. Otherwise the output deviates as minimally as possible from the nominal controller. 
The solution for the QP can be obtained in a closed-form \cite{alan2022control}. For a single input system, ${u\in\R}$, this solution simplifies to:
\begin{empheq}[left={u^*(x)=\empheqlbrace}]{align}
    \max\{u_{\rm nom}(x), u_{\rm safe}(x)\} \quad & \textrm{if}~ {L_gb(x) > 0},  \\
    \min\{u_{\rm nom}(x), u_{\rm safe}(x)\} \quad & \textrm{if}~ {L_gb(x) < 0}, \label{eq:app_safetyfilter_singleinput}  \\
    u_{\rm nom}(x) \quad & \textrm{if}~ {L_gb(x) = 0},
\end{empheq}
where 
\begin{equation}
\label{eq:app_usafe}
    u_{\rm safe}(x)=-\frac{L_fb(x)+\alpha(b(x))}{L_gb(x)},
\end{equation}
and terms ${L_fb(x)=\nabla b(x) \cdot f(x)}$ and ${L_gb(x)=\nabla b(x) \cdot g(x)}$ denote Lie derivatives. 

Considering model \eqref{eq:carfollowingmodel} and a CBF in the form \eqref{eq:safeset}, that corresponds to \eqref{eq:app_C} with ${x=[h,v,v_1]^\top}$ and ${b(x)=h-\rho(v,v_1)}$, and using Assumption~\ref{ass:rho}, one may obtain 
\begin{equation}
    L_gb(x) < 0,
\end{equation}
which suggests that the safety filter \eqref{eq:app_safetyfilter_singleinput} is applicable for the car-following scenario; cf.~\eqref{eq:safeintegration}.

\bibliographystyle{IEEEtran}
%\bibliography{_Aux/Alan_bib}

\vspace{-5 mm}
\begin{IEEEbiography}[{\includegraphics[width=1in,height=1.25in,clip,keepaspectratio]{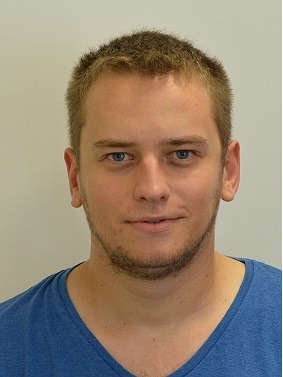}}]{Anil~Alan} received the BSc degree in mechanical engineering from Middle East Technical University, Turkey, in 2012 and the MSc degree in Bilkent University, Turkey, in 2017. He is currently pursuing the PhD degree in mechanical engineering with the University of Michigan, Ann Arbor, MI, USA. His current research interests include control of connected autonomous vehicles, robust safety-critical control, and vehicle dynamics. 
\end{IEEEbiography}

\vspace{-5 mm}
\begin{IEEEbiography}[{\includegraphics[width=1in,height=1.25in,clip,keepaspectratio]{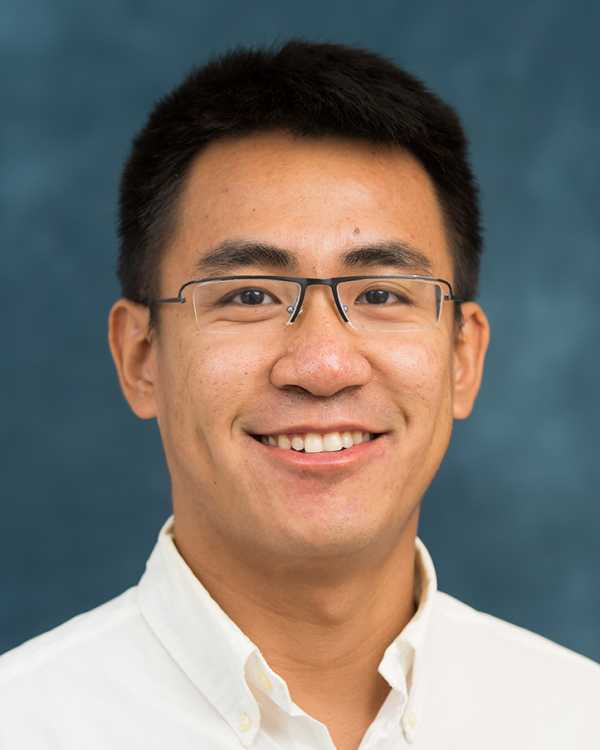}}]{Chaozhe~R.~He} received the BSc degree in applied mathematics from the Beijing University of Aeronautics and Astronautics in 2012, the MSc and PhD in Mechanical Engineering from the University of Michigan, Ann Arbor, USA, in 2015 and 2018 respectively. Dr.~He is with Plus.ai Inc.\ and is working on planning and control algorithm development. His research interests include dynamics and control of connected automated vehicles, optimal and nonlinear control theory, and data-driven control.
\end{IEEEbiography}

\vspace{-5 mm}
\begin{IEEEbiography}[{\includegraphics[width=1in,height=1.25in,clip,keepaspectratio]{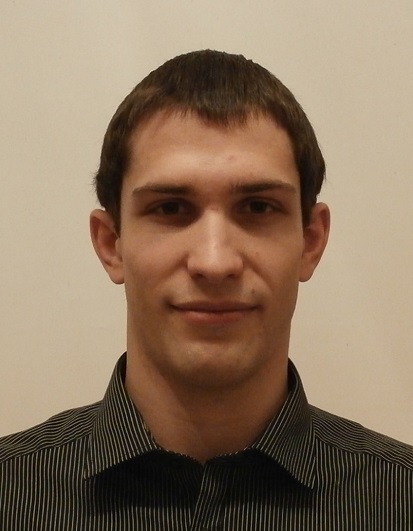}}]
{Tamas G. Molnar} received his BSc degree in Mechatronics Engineering, MSc and PhD degrees in Mechanical Engineering from the Budapest University of Technology and Economics, Hungary, in 2013, 2015 and 2018.
He held postdoctoral position at the University of Michigan, Ann Arbor between 2018 and 2020.
Since 2020 he is a postdoctoral fellow at the California Institute of Technology, Pasadena.
His research interests include nonlinear dynamics and control, safety-critical control, and time delay systems with applications to connected automated vehicles, robotic systems, and machine tool vibrations.
\end{IEEEbiography}

\vspace{-5 mm}
\begin{IEEEbiography}[{\includegraphics[width=1in,height=1.25in,clip,keepaspectratio]{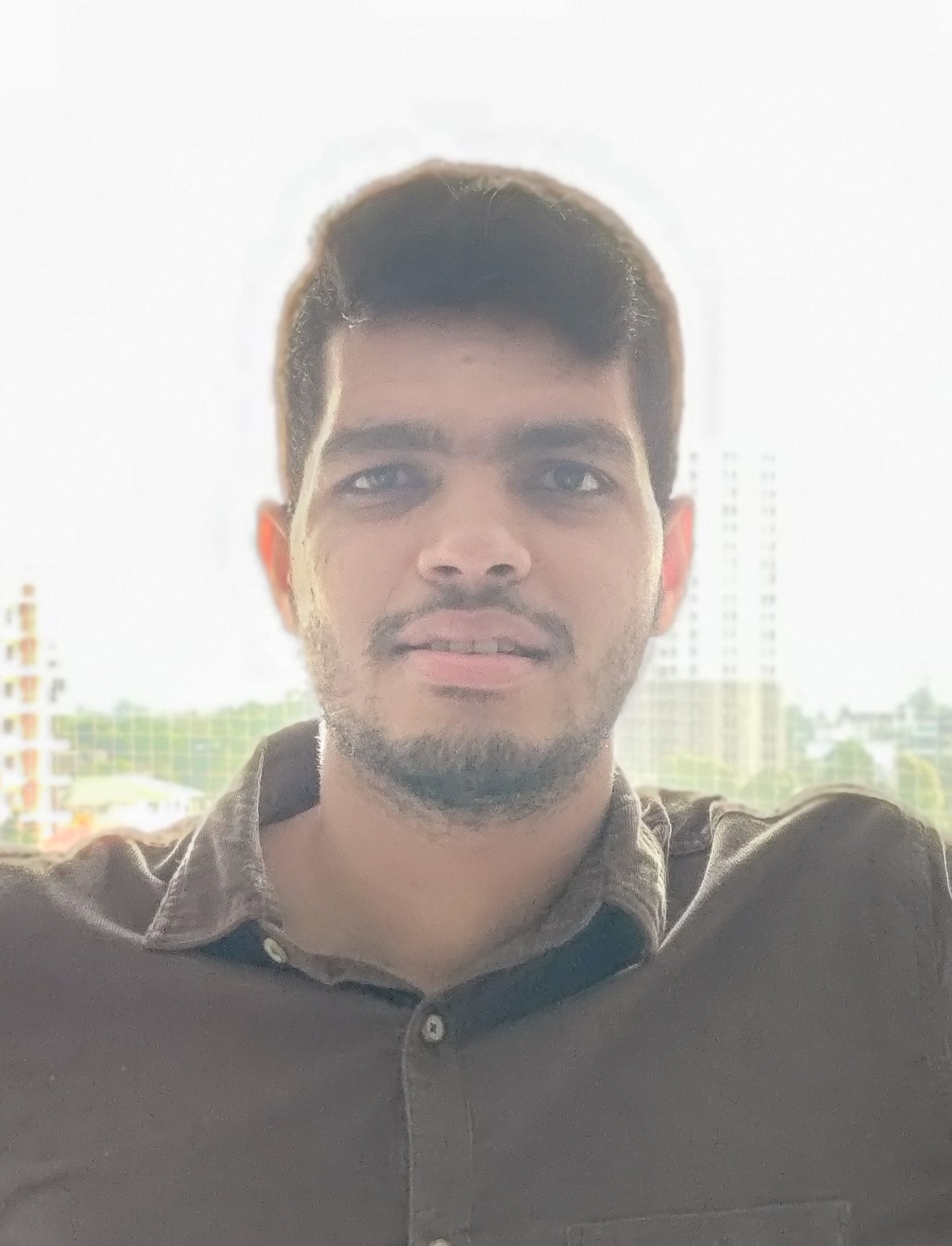}}]
{Johaan Chacko Mathew} received a BTech in mechanical engineering from the Indian Institute of Technology Madras, Chennai, India, in 2018. He received an MS in mechanical engineering from the University of Michigan, Ann Arbor, USA, in 2020. He is with Visteon Corporation as part of their motion controls and execution team in their advanced driver-assistance (ADAS) program. His research interests include system modeling, identification, and control of dynamical systems, mechatronic systems, and vehicle dynamics.
\end{IEEEbiography}

\vspace{-5 mm}
\begin{IEEEbiography}
[{\includegraphics[width=1in,height=1.25in,clip,keepaspectratio]{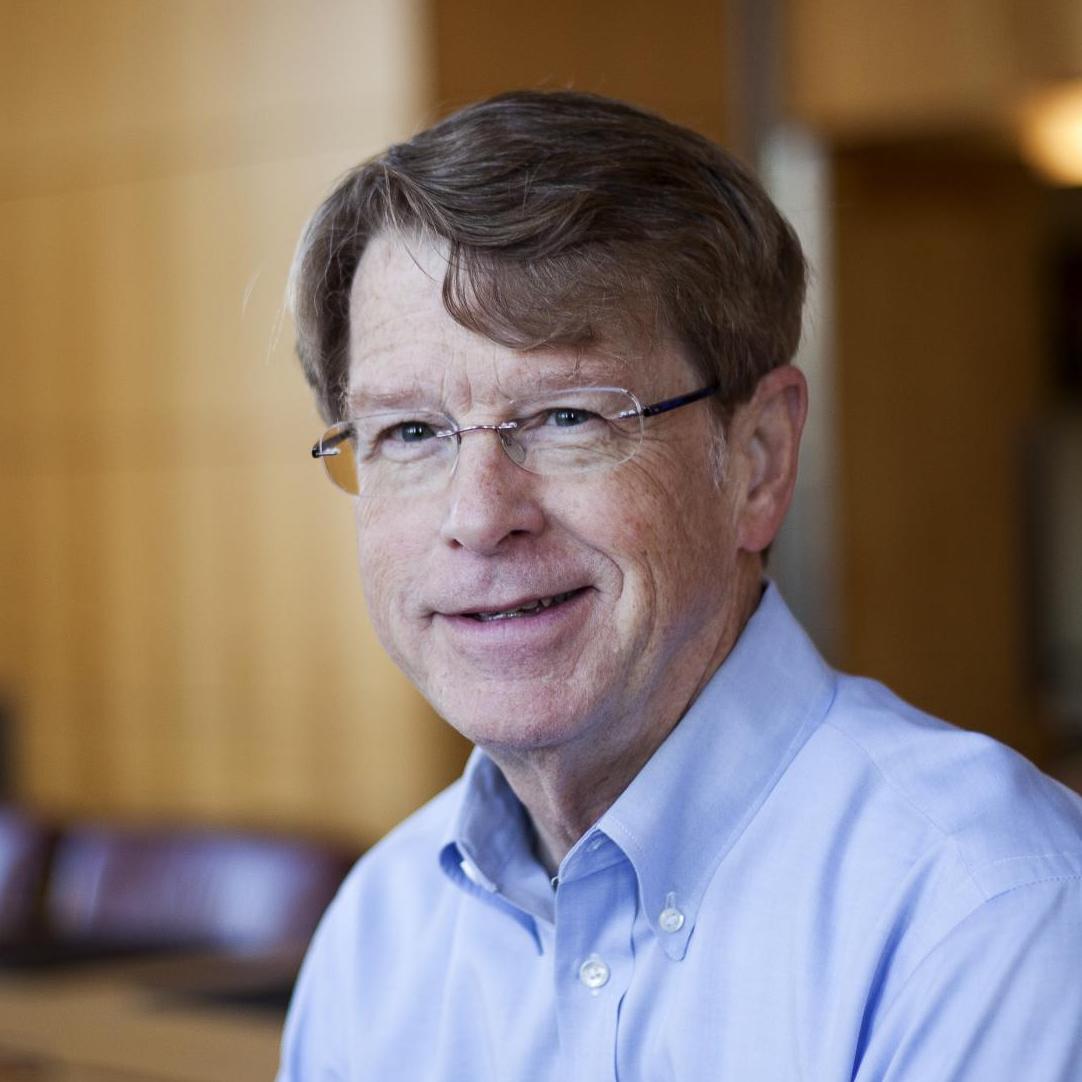}}]
{A. Harvey Bell} is a Professor of Engineering Practice at the University of Michigan, Ann Arbor and he also serves as the Co-director of the Multidisciplinary Design Program. He spent his 39-year long career in the automotive industry with General Motors where some of his significant achievements were: Chief Engineering a 2.5 Liter Engine, Vehicle Chief Engineering the 4th Generation Camaro and Firebird, Executive Director of the Advanced Vehicle Development Center for North America. He is a graduate of the University of Michigan and undertook graduate studies at the University of Pennsylvania and University of Michigan, Dearborn. 
\end{IEEEbiography}

\vspace{-5 mm}
\begin{IEEEbiography}
[{\includegraphics[width=1in,height=1.25in,clip,keepaspectratio]{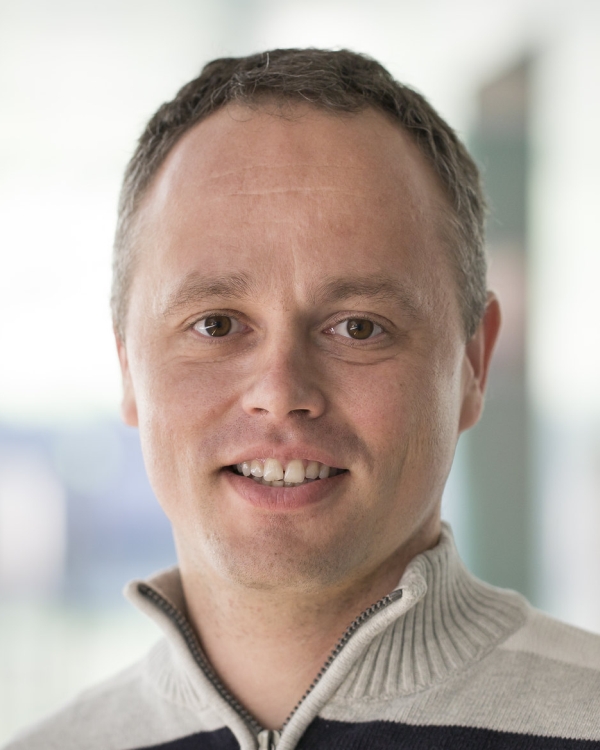}}]
{G{\'{a}}bor Orosz} received the MSc degree in Engineering Physics from the Budapest University of Technology, Hungary, in 2002 and the PhD degree in Engineering Mathematics from University of Bristol, UK, in 2006. He held postdoctoral positions at the University of Exeter, UK, and at the University of California, Santa Barbara. In 2010, he joined the University of Michigan, Ann Arbor where he is currently an Associate Professor in Mechanical Engineering and in Civil and Environmental Engineering. His research interests include nonlinear dynamics and control, time delay systems, and machine learning with applications to connected automated vehicles, traffic flow, and biological networks.
\end{IEEEbiography}

\end{document}